# The conjugated null space method of blind PSF estimation and deconvolution optimization

**Yuriy A Bunyak**[1]**, Roman N Kvetnyy**[2] **and Olga Yu Sofina**[2]

[1]InnoVinn Inc. Vinnitsa, Ukraine
[2]Vinnitsa National Technical University, Vinnitsa, Ukraine

E-mail: yuri.bunyak@innovinn.com

**Abstract.** We have shown that the vector of the point spread function (PSF) lexicographical presentation belongs to the left side conjugated null space (NS) of the autoregression (AR) matrix operator on condition the AR parameters are common for original and blurred images. The method of the PSF and inverse PSF (IPSF) evaluation in the basis of the NS eigenfunctions is offered. The optimization of the PSF and IPSF shape with the aim of fluctuation elimination is considered in NS spectral domain and image space domain. The function of surface area was used as the regularization functional. Two methods of original image estimate optimization were designed basing on maximum entropy generalization of sought and blurred images conditional probability density and regularization. The first method uses balanced variations of convolutions with the PSF and IPSF to obtaining iterative schema of image optimization. The variations balance is providing by dynamic regularization basing on condition of the iteration process convergence. The regularization has dynamic character because depends on current and previous image estimate variations. The second method implements the regularization of the deconvolution optimization in curved space with metric defined on image estimate surface. The given iterative schemas have fast convergence and therefore can be used for reconstruction of high resolution images series in real time. The NS can be used for design of denoising bilateral linear filter which does not introduce image smoothing.

## 1. Introduction

Many modern applications need real time reconstruction of high resolution images of some millions pixels size, which are corrupted by defocusing, medium penetration, camera jitter and other factors. Usually, the model of corruption is presented as convolution of original image signal and point spread function (PSF) [5, 26, 30]. The exact PSF shape is unknown in majority cases. So, the image reconstruction problem includes the blind identification of the PSF and deconvolution. The both problems are ill-posed and their solution is approximate in accordance with chosen optimization criterion.

There are known the methods of the blind PSF estimation using only degraded image. These methods are primary statistical [2, 5, 6, 9, 14, 27, 35, 36, 37, 42, 55] or image model based [12, 31, 41]. The best results where given when an image has sparsity presentation and blur can be localized, for example [9, 36], or eliminated by spectral transforms [16]. But natural images look primary as composition of textures which do not have such presentation.

The principal possibility of the PSF estimation problem solution analytically as an algebraic problem was proven by Lane and Bates [33]. They have shown that convolved components of any multidimensional signal can be separated on condition the signal dimension is greater than one. The zeros of the Z-transform (ZT) of multi dimensional components create a hypersurface which can be separated and zeros of the individual components can be recognized up to a complex scale factors. The general approach to zero-sheet separation is based on finding of mutually independent zeros groups which create singularity of Vandermonde like matrix [1, 30, 33]. Such solution is equivalent to finding of a null space (NS) of mentioned matrix. The zero-sheet separation approach is very complicated for implementation because zeros accuracy evaluation depends on resolution of image spectral presentation with the help of a discrete samples set which is limited by available combinations number of spectral components and should not be enormously large.

Model of image signal can be used instead of the image directly to reduce the problem complexity. As it was shown in [6, 31], the autoregressive and moving average (MA) model enables to separate original image and PSF as parts which correspond to the AR and MA models accordingly. The AR model characteristic polynomial contains zeros that are equal to ZT zeros. But this zeros set includes original image zeros as well as PSF zeros because the AR model parameters can be estimated only by means of measured degraded image. An elimination of the PSF zeros influence can be reached by using some number of images which was gotten as different degradations of the same original image. As it was shown in [45,46], if there are at least three differently degraded images then the original image can be reconstructed exactly as the right side NS of the operator which acts as convolution of the AR parameters and image data vectors. Size of the vector of the lexicographical presentation of the original image signal matrix is equal to NS size and so the NS approach is appropriate to relatively small images reconstruction for a time far from usual period of image frames storing by camera.

When the PSF is known, the problem of deblurring can be considered as inverse filtration of observed image. Methods of inverse filters implementation are direct, mainly in spectral domain, and indirect, with using maximum likelihood, Bayesian and variational approaches. Wiener spectral method [5, 30] lies in base of majority methods of inverse filtration [12, 41, 44]. The designed methods are intended for regularization of spectrum inversion and optimization of the PSF spectrum shape with the aim of elimination of a noise influence.

There are some approaches to simultaneous iterative deconvolution and PSF estimation. The Lucy-Richardson (LR) maximum likelihood (ML) method [38, 48] of deconvolution was supplemented in [24] by iterative schema of the PSF estimation and was developed in [7, 20, 25, 47, 52] in extended manner. LR method has some limitations, such as slow convergence, sensitivity to noise, image artefacts presence. The methods of variational optimization [13, 14, 15, 23, 42, 55], statistical Bayesian and maximum a posterior models of blur [2, 9, 35, 36, 37, 41, 43, 51] are using to obtain iterative schemas of sought image and PSF estimates. Limitations and advantages of the variational method depend on estimated prior model of the PSF and chosen regularization operator. In the case of large image the variational method implementation is difficult because it uses matrix operators which rows size equal to processed image lexicographical presentation size. The parametrization of the least square, Bayesian or Euler-Lagrange (EL) optimization problems in accordance with gradient methods gives appropriate schemas for iterative optimization of original image estimate [4, 18, 22, 27, 32, 34, 49, 54]. These schemas are not always stable and convergent, they usually require hundreds of iterations for attainment of a convergence limit condition because of slow relaxation.

We consider the problem of arbitrary size images series reconstruction from the point of view of their real time processing. This aim means that original image estimate can be found by one step deconvolution using the inverse point spread function (IPSF) and it is allowable to use some optimization iterations. It is assumed that blur characteristic changing slowly and the IPSF can be estimated without real time restriction. Therefore, the problem solution includes the PSF estimation, definition of the optimal IPSF and evaluation of the primary image estimate, optimization of the given estimate by some number of iterations.

## 2. The conjugated NS (CNS) method of PSF estimation

We consider the generalized approach to blind estimation of convolved components of a degraded by blur image signal basing on the NS model. The model is created by the linear shift invariant singular operator. The relation of the operator and convolved components spectrum in the Z-domain shows that the right side NS and the left side NS – CNS, of the operator associate with original image and the PSF. The operator presentation in the left side eigenfunctions basis allows create sought PSF and IPSF shapes basing on the spectral presentation of the image first and second moments. The discrete presentation of the operator gives the algorithm of the PSF estimation.

*2.1. The problem formulation and solution in image model operator CNS*

The linear model of a degraded measured image $X = X(x, y) \in \Omega$, where $\Omega$ is an image space, $x, y$ – image pixels coordinates, in noise free case has the form of convolution

$$X(\varsigma,\xi) \cong \int_\Xi H(\theta,\tau)S(\varsigma-\theta,\xi-\tau)d\theta d\tau = H*S \qquad (1)$$

of the blur characteristic PSF $H \in \Xi \subset \Omega$ and original image $S \in \Omega$ [5, 30]. The size of the PSF space $\Xi$ is far less than the image space $\Omega$. It is required to obtain the inverse expression

$$S(\varsigma,\xi) \cong \int_\Xi G(\theta,\tau)X(\varsigma-\theta,\xi-\tau)d\theta d\tau = G*X \qquad (2)$$

for image reconstruction on condition that the blur function $H$ is unknown.

One of the ways of the problem solution is based on the assumption that original and degraded images possess algebraic structure which is invariant to blur operator (1) action. Such common algebraic structure may be defined as operator A which NS is constituted by the image matrices $X$ and $S$:

$$\int_\Omega A(x,y;\varsigma,\xi)S(\varsigma,\xi)d\varsigma d\xi \cong 0; \qquad (3)$$

$$\int_\Omega A(x,y;\varsigma,\xi)X(\varsigma,\xi)d\varsigma d\xi \cong 0. \qquad (4)$$

The integral expressions (3), (4) are linear and therefore degradation transform (1) and inverse transform (2) do not effect on them. In general, the exact NS does not exist and so we used approximate equality in (3) and (4). Also, expressions (1) and (2) are approximate models of real processes. Expression (4) can be rewritten with the account of convolution (1) as

$$\int_\Omega A(x,y;\varsigma,\xi)\int_\Xi H(\theta,\tau)S(\varsigma-\theta,\xi-\tau)d\theta d\tau d\varsigma d\xi =$$
$$\int_\Xi H'(\theta,\tau)\int_\Omega A(x,y;\varsigma,\xi)S(\varsigma-\theta,\xi-\tau)d\theta d\tau d\varsigma d\xi \cong 0, \qquad (5)$$

where $\cdot'$ – the conjugation in respect to the integration sequence change. If the model operator is shift invariant: $A(x,y;\varsigma,\xi) = A(x-\varsigma, y-\xi)$, then equation (5) is satisfied for arbitrary $H$ because the integral about $\Omega$ is equal to integral (3).

Let $W(x,y;\varsigma,\xi) = \delta(x-\varsigma)\delta(y-\xi) \subset \Xi \times \Omega$ is 2D $\delta$-function window. The PSF can be presented in the image space $\Omega$ as

$$H_w(\varsigma,\xi) = \int_\Xi H(\theta,\tau)W(\varsigma-\theta,\xi-\tau)d\theta d\tau. \qquad (6)$$

Then expressions (1) and (5) are equivalent to the following ones:

$$X(\varsigma,\xi) = \int_\Omega H_w(\varsigma-\theta,\xi-\tau)S(\theta,\tau)d\theta d\tau;$$
$$\int_\Omega A(x-\varsigma, y-\xi)\int_\Omega H_w(\varsigma-\theta,\xi-\tau)S(\theta,\tau)d\theta d\tau d\varsigma d\xi \cong 0.$$

It is evident that the functions $H_w$ and $S$ are at the symmetric position in the equations of the problem statement (1), (3) and (4). Therefore the presumption that

$$\int_\Omega A(x-\varsigma, y-\xi)H_w(\varsigma,\xi)d\varsigma d\xi \cong 0 \qquad (7)$$

can be made. The presumption argumentum is basing on the presentation of the shift invariant algebraic structure $A(x-\varsigma, y-\xi)$ in the Z-space as a polynomial $A(Z)$ which roots create a zero hypersurface $Z_A : A(Z_A) \cong 0$. The expressions (3), (4) and (7) can be considering as whitening filters [40]. Such filter is whitening in respect to any signal which zero hypersurface $Z_{any} : Z_{any} \subset Z_A$. The equation (1) and the conditions (3) and (4) mean that zero hypersurfaces corresponding to $X$, $H_w$ and $S$ are bounded by the spectral relations:

$$X(Z_X) = H_w(Z_{Hw}) \cdot S(Z_S) ; \; Z_X : Z_X \subseteq Z_A ; \; Z_S : Z_S \subset Z_X \cap Z_A ;$$

$$Z_{Hw} : Z_{Hw} \subset Z_X \cap Z_A ; \; Z_S \cup Z_{H_w} = Z_X \subseteq Z_A . \tag{8}$$

The spectral relations (8) select the function $H_w$ up to a scale factor which satisfies the equations (5) and (7) as well as original image $S$ in (3).

As it follows from equations (5) and (6), equation (7) can be rewritten as

$$\int_\Xi H'(\theta,\tau) A(\varsigma-\theta, \xi-\tau) d\theta d\tau \cong 0 . \tag{9}$$

So, we can made the following conclusion.

If image signal $X \in \Omega$ is a convolution of two components – $S \in \Omega$ and $H \in \Xi \subset \Omega$, and it belongs to NS $N_\Omega(A)$ of the linear shift invariant singular operator $A \in \Omega \times \Xi$, then the convolving components belong to NS $N_\Omega(A)$ and conjugated NS $N_\Xi(A)$: $S \in N_\Omega(A)$ and $H \in N_\Xi(A)$. The operator full NS $N(A) = N_\Omega(A) \times N_\Xi(A)$. The conjugation means an agreement of the algebraic structure action on functions in mentioned spaces.

This conclusion has only sufficient proof by sentences (7) and (9). The proof of necessity is not possible because of NS specific. If convolution (5) is formed by two functions and one of them belonging to corresponding NS's then the second function can be arbitrary. Additional conditions are needed for exact definition of $H(x, y)$ in (9). Spectral relations (8) can serves as such conditions. But function $H(x, y)$ can be defined explicitly only if the zero hypersurface $Z_A : A(Z_A) \cong 0$ roots are evaluated or if the zero hypersurface $Z_H : A(Z_H) \cong 0$ roots are given, for example, by searching [1]. Then $H(x, y)$ may be found using Fourier transform. We will consider another approach to founding the PSF in (9) using spectral domain.

Let define the operator

$$B(\varsigma-\theta, \xi-\tau) = \int_\Omega A(x-\theta, y-\tau) A(x-\varsigma, y-\xi) dxdy \tag{10}$$

in the space $\Xi \times \Xi$. This operator is symmetrical and therefore it has orthogonal system of eigenfunctions $V_k(\varsigma, \xi) \in \Xi$:

$$\int_\Xi V_k(\theta, \tau) B(\varsigma-\theta, \xi-\tau) d\theta d\tau = \lambda_k V_k(\varsigma, \xi) , \tag{11}$$

where the operator $B$ eigen space (ES) eigenvalues $|\lambda_k| \gg 0$ and the operator NS eigenvalues $\lambda_k \approx 0$, $k = 1, 2, ... \in ES \cup NS$. The operator $B$ is a linear superposition of the operators $A$, so it has the same rank that the operator $A$ in (9). Also, the operator $A$ left side eigenfunctions are the same that in (10), or its CNS is equal to the operator (10) NS. This is evident from the following sentence:

$$\int_\Omega \int_\Xi V_k(\theta, \tau) A(x-\theta, y-\tau) d\theta d\tau A(x-\varsigma, y-\xi) dxdy =$$
$$\sqrt{\lambda_k} \int_\Omega V_k(\theta, \tau) W(x-\theta, y-\tau) A(x-\varsigma, y-\xi) dxdy = \lambda_k V_k(\varsigma, \xi) .$$

The sought PSF $H(x, y)$ should converge to the trivial PSF

$$H(x, y) \to \delta(x - x_0, y - y_0) \cong \sum_{k \in NS} V_k(x, y) V_k(x_0, y_0) : \forall (x_0, y_0) \in \Xi \qquad (12)$$

on the condition $X \to S$. The expression (12) becomes equality when the sum is complemented by components corresponding to $k \in ES$ or, in limit, when all eigenvalues in (11) vanish, so the model operator vanishes too. This condition corresponds to the trivial case $X = S$. Consequently, we should find the operator of the problem statement (3) and (4) with account of difference between initial $X$ and original $S$ images. This difference shows image signal fronts changes in result of blur action. We can use initial image gradient $\nabla X(x, y) = (\partial/\partial x + \partial/\partial y) X(x, y)$ which shows original image signal changes with account of blur action instead of the unknown difference $X - S$. The problem consists in a separation of the blur part of the changes which belong to the model operator CNS in accordance with (9).

The operator (10) has well known spectral presentation [53] as

$$B(\varsigma - \theta, \xi - \tau) = \sum_{k \in ES} \lambda_k V_k(\varsigma, \xi) V_k(\theta, \tau). \qquad (13)$$

The expression (13) shows the dynamic feature of the model operator which is given by the shift transform in the eigenfunctions basis. As it follows from (8) and (9), the sought PSF has the same dynamic feature and it can be presented by analogous expression in the extended space $\Xi \times \Xi$ as the series

$$H(\varsigma - \theta, \xi - \tau) = \sum_{k \in NS} v_k V_k(\varsigma, \xi) V_k(\theta, \tau), \qquad (14)$$

where the coefficients vector $\mathbf{v} = [v_k]_{k \in NS}$ creates the PSF estimate (13) shape. The orthogonality of the ES and CNS in (13) and (14) provides equation (9) fulfillment. The projection of the extended PSF (14) into $\Xi$ with account of the normalization condition (12) can be found as the modified $\delta$-function (12):

$$H(x, y) \cong \sum_{k \in NS} v_k V_k(x, y) V_k(x, y), \qquad (15)$$

The coefficients vector $\mathbf{v}$ can be estimated by the help of first and second moments of the image gradient.

The 2D auto-correlation function

$$R_H(\theta, \varsigma; \tau, \xi) = \int_\Xi H(x - \theta, y - \tau) H(x - \varsigma, y - \xi) dx dy \qquad (16)$$

in the basis of eigenfunctions has the diagonal spectrum

$$\iint_\Xi R_H(\theta, \varsigma; \tau, \xi) V_i(\theta, \varsigma) V_k(\tau, \xi) d\theta d\tau d\varsigma d\xi = \rho_{H_k} \delta_{i,k} \qquad (17)$$

because of the spectral presentation (14). Let consider the same auto-correlation function $R_X(\theta, \varsigma; \tau, \xi)$ of image gradient. The diagonal of the function $R_X(\theta, \tau; \theta, \tau)$ characterizes power distribution of $\nabla X$ shifted fragments. The cross-diagonal $R_X(\theta, -\tau; \theta, -\tau)$ characterizes power distribution of mutually shifted fragments. The cross-diagonal shows image changes at the shift and therefore it maps blur feature too. We can define cross-auto-correlation function in respect to function like (16) for $\nabla X$ as the following one –

$$R_{crX}(\theta, \varsigma; \tau, \xi) = \int_\Omega \nabla X(x - \theta, y - \tau) \nabla X(x + \varsigma, y + \xi) dx dy \in \Xi \times \Xi, \qquad (18)$$

and create the spectral filter for separation of the PSF part in the image changes as the diagonal spectrum in the CNS basis:

$$\rho_{H_k} \sim \rho_{X_{k,k}} = \iint_\Xi R_{crX}(\theta,\varsigma;\tau,\xi) V_k(\theta,\varsigma) V_k(\tau,\xi) d\theta d\tau d\varsigma d\xi : \quad \forall k \in NS. \qquad (19)$$

The estimate (19) is basing on the assumptions that the projection of the function (18) in the CNS maps the correlation function (16) with its feature of shift transform given by (14) and (17), also, the correlation of image signal changes is presented by non-diagonal components $\rho_{crX_{i,k}}$ and components in the ES basis which is orthogonal to the CNS basis. The vector **v** elements estimate in (14) and (15) looks as

$$v_k^2 \propto \rho_{X_{k,k}} : \forall k \in NS. \qquad (20)$$

The estimate (20) corresponds to variational part of the PSF. The PSF values level we can estimate using the averaged image gradient mutually shifted fragments

$$\Omega^{-1} \int_\Omega \nabla X(x+\varsigma-\theta, y+\xi-\tau) dx dy = \nabla \overline{X}(\varsigma-\theta, \xi-\tau) \in \Xi \times \Xi. \qquad (21)$$

The function (21) can be presented with account of (1), (14) and convolution symmetry in respect to convolved components in the following manner:

$$\nabla \overline{X}(\varsigma-\theta, \xi-\tau) = \int_\Xi H(\varsigma-x, \xi-y) \nabla \overline{S}(x-\theta, y-\tau) dx dy =$$

$$= \sum_{k \in NS} v_k \int_\Xi V_k(\varsigma,\xi) V_k(x,y) \nabla \overline{S}(x-\theta, y-\tau) dx dy \qquad (22)$$

where $\nabla \overline{S}$ is the original image $S$ gradient averaged fragments like (21). Fourier transform gives the averaged spectral matrix with elements

$$\overline{\omega}_{X_{i,j}} = \iint_\Xi V_i(\varsigma,\xi) V_j(\theta,\tau) \nabla \overline{X}(\varsigma-\theta, \xi-\tau) d\varsigma d\xi d\theta d\tau. \qquad (23)$$

The similar matrix with the elements $\overline{\omega}_{S_{i,j}}$ can be evaluated for $\nabla \overline{S}$ too. In accordance with (15) and (17), only diagonal elements of these matrices are interest for us. The substitution of the second expression of (22) into (23) yields the following expression for the diagonal elements –

$$\overline{\omega}_{X_{k,k}} = v_k \overline{\omega}_{S_{k,k}}. \qquad (24)$$

As it follows from (24), at $X \to S$ the coefficients $v_k \to 1$ in accordance with condition (12). We can suppose that blur is invariant to averaging procedure (21) and the spectrum of the averaged original image frames $\overline{\omega}_{S_{k,k}} \approx \overline{\omega}_S + \eta_{k,k} : \forall k \in NS$, where constant $\overline{\omega}_S \gg \eta_{k,k}$. Then $v_k$ in (24) can be evaluated up to scale factor $\overline{\omega}_S$. The square of this scale factor is present in (19) too. So, the joined with (20) estimate of the PSF spectrum coefficients may be found as

$$v_k \cong \overline{\omega}_S^{-1} sign(\rho_{X_{k,k}} + \overline{\omega}_{X_{k,k}}^2) \sqrt{\left| \rho_{X_{k,k}} + \overline{\omega}_{X_{k,k}}^2 \right|}. \qquad (25)$$

The scale factor in (25) can be eliminated by the normalization of the PSF:

$$\int_\Xi H(x,y) dx dy = 1. \qquad (26)$$

The equations (1) and (2) mean that symmetric convolution

$$\int_\Xi G(x-\varsigma, y-\xi)H(\theta-\varsigma,\tau-\xi)d\varsigma d\xi \cong \delta(x-\theta, y-\tau). \tag{27}$$

We can rearrange by places the images $X$ and $S$, blur functions $H$ and $G$ in expressions (1)-(4), (7) symmetrically. This has shown that the inverse PSF $G$ belongs to the same NS that the PSF and it can be defined as the series (14) or (15), for example

$$G(x-\varsigma, y-\xi) \cong \sum_{k\in NS} u_k V_\kappa(x,y)V_k(\varsigma,\xi). \tag{28}$$

The substitution of expressions (12), (14) and (28) into (27) and integration over variables yields that $u_k \propto v_k^{-1}$. This assessment is approximate because it was got with using only $\delta$-function projection (12) into NS. The unsymmetrical convolution

$$\int_\Xi G(\varsigma,\xi)H(\theta-\varsigma,\tau-\xi)d\varsigma d\xi \cong \delta(\theta,\tau) \tag{29}$$

with substitution (28) gives the expression for the vector $\mathbf{u}=[u_k]_{k\in NS}$ evaluation with account of $\delta$-function completely:

$$\sum_{k\in NS} u_k \int_\Xi V_k(\varsigma,\xi)V_k(\varsigma,\xi)H(\theta-\varsigma,\tau-\xi)d\varsigma d\xi \cong \delta(\theta,\tau). \tag{30}$$

We have offered the CNS method of the PSF and IPSF estimation. The method includes the problem formulation – expressions (1) and (2), main supposition to the image blur model creation – (3)-(7), the problem (7) solution in the NS eigenfunctions basis (11) as the series (14) and (28). The next problem is the method digital implementation with PSF, IPSF and estimated image optimization.

*2.2. The null space method of image deblurring*

The presented above generalized schema was obtained on the base of the NS method [45] of original image reconstruction using $K \geq 3$ degraded by different filters images. The NS method includes four following steps.

At the first step the extended matrix is created

$$\mathbf{X} = [\mathbf{x}_{i+k,l+m}]_{i=1...P,\ l=1...Q}^{k=0...N_x-P-1,\ m=0...N_y-Q-1} \tag{31}$$

where low indices point on image matrix columns and high indices point on rows, $N_x \times N_y$ – image matrices $X^1...X^K$ size, $P$, $Q$ – the NS model parameters, $\mathbf{x}_{i,k} = [\mathbf{x}_{i,k}^1...\mathbf{x}_{i,k}^K]^T$ – the images pixels vectors with indices $i$ and $k$ along coordinates $x$ and $y$, $T$ – the transposition.

The second step – the NS vector $\mathbf{a} = \mathbf{vec}([a_{i,k}]_{i=1...P}^{k=1..Q})$, the lexicographical presentation of the matrix in the parenthesis, has to be found:

$$\mathbf{X} \cdot \mathbf{a} \cong \mathbf{0} \tag{32}$$

At the third step the matrix is formed.

$$\mathbf{A} = \begin{bmatrix} \mathbf{A}_1\mathbf{A}_2...\mathbf{A}_P\mathbf{0}\ ...\mathbf{0} \\ \mathbf{0}\ \mathbf{A}_1\mathbf{A}_2...\mathbf{A}_P...\mathbf{0} \\ \vdots\ \ \ \ \ \ \ \ \ \ \ \ \ \ \ \ \ \ddots\ \ \ \ \ \vdots \\ \mathbf{0}\ ...\mathbf{0}\ \mathbf{A}_1\mathbf{A}_2...\mathbf{A}_P \end{bmatrix}; \ \mathbf{A}_k = \begin{bmatrix} a_{k,1}a_{k,2}...a_{k,Q}0\ ...0 \\ 0\ a_{k,1}a_{k,2}...a_{k,Q}...0 \\ \vdots\ \ \ \ \ \ \ \ \ \ \ \ \ \ \ \ \ \ddots\ \ \ \ \ \vdots \\ 0\ ...0\ a_{k,1}a_{k,2}...a_{k,Q} \end{bmatrix}, \qquad (33)$$

where $\mathbf{0}$ – the null matrix. The matrix (33) includes $N_x \cdot N_y$ columns, minimal number of rows is $P \cdot Q$.

The fourth step, NS vector $\mathbf{s}$ has to be found:

$$\mathbf{A} \cdot \mathbf{s} \cong \mathbf{0}. \qquad (34)$$

The vector

$$\mathbf{s} = \mathbf{vec}(S) \qquad (35)$$

is the lexicographical presentation of the sought original image $S$ or its estimate.

As it follows from expressions (31) – (33), the equations (32) can be rewritten as

$$\mathbf{A} \cdot \mathbf{x} \cong \mathbf{0}, \qquad (36)$$

where $\mathbf{x} = \mathbf{vec}(X)$ and $X$ is any of $K$ given images. The equations (34) and (36) are discrete presentation of the equations (3) and (4) correspondingly. The matrix (33) is discrete presentation of the linear shift invariant operator of the algebraic structure in (3) and (4).

*2.3. Null space of PSF*

We can rewrite the expressions (27) and (31) as the equations system

$$\sum_{i=1}^{P} \sum_{k=1}^{Q} \mathbf{x}_{i+n,k+m} a_{i,k} \cong 0, \qquad (37)$$

where $n = 0...N_x - P$, $m = 0...N_y - Q$, which looks as 2D AR model. However; this is not quietly AR model because expression (37) is not intended for autoregressive process forming in the manner of linear prediction or whitening filter [40]. Expression (37) is the polynomial model of an image elements mutual relation. With account that blur is a diffusion process around of each image point, model (37) should be centralized – its central element $a_{p,q} \equiv 1$.

As it was shown in [45], matrix $\mathbf{X}$ (31) and analogous extended original image matrix $\mathbf{S}$ have same NS, or in other words same AR model (37), on condition that matrix (31) includes $K \geq 3$ image matrices obtained by degradation filters with coprime characteristic polynomials. The matrix (33) rows size is equal to vector (35) size $N_x \cdot N_y$ and the model parameters are the following: $2P - 1 = N_x$; $2Q - 1 = N_y$. So, the AR model of very high order was used in the NS method [45] for original image reconstruction and therefore it needs such restriction. The assumption that degraded and original images posses the same AR model was used in [6, 31] too. In the last cases one image was used for model evaluation without any restrictions on a degradation filter characteristic and an appropriate model order.

As the image (1) is the convolution of the original image and PSF, the model (37) is common in respect to these components. But there is one sufficient difference, the images $X, S \in \Omega$, the PSF $H = [h_{i,k}]_{i=1...L}^{k=1..M} \in \Xi \subset \Omega$, where $L \times M$ is the PSF subspace size. Let's define the convolution matrix operator in the PSF subspace $\Xi$.

$$\mathbf{H} = \begin{bmatrix} \mathbf{H}_1\mathbf{H}_2...\mathbf{H}_L\mathbf{0}\ ...\mathbf{0} \\ \mathbf{0}\ \mathbf{H}_1\mathbf{H}_2...\mathbf{H}_L...\mathbf{0} \\ \vdots\ \ \ \ \ \ \ \ \ \ \ \ \ \ \ \ \ \ddots\ \ \ \ \ \vdots \\ \mathbf{0}\ ...\mathbf{0}\ \mathbf{H}_1\mathbf{H}_2...\mathbf{H}_L \end{bmatrix}; \ \mathbf{H}_k = \begin{bmatrix} h_{k,1}h_{k,2}...h_{k,M}0\ ...0 \\ 0\ h_{k,1}h_{k,2}...h_{k,M}...0 \\ \vdots\ \ \ \ \ \ \ \ \ \ \ \ \ \ \ \ \ \ddots\ \ \ \ \ \vdots \\ 0\ ...0\ h_{k,1}h_{k,2}...h_{k,M} \end{bmatrix}. \qquad (38)$$

The operator (38) size is equal to $(L \cdot M) \times (2L-1)(2M-1)$ and it can be enlarged by zeros inserting in to columns and mutually shifted rows addition up to $(2P-1)(2Q-1) \times N_x N_y$.

The equation (36) can be rewritten in two mutually equivalent forms as

$$\mathbf{A} \cdot \tilde{\mathbf{H}} \cdot \mathbf{s} \cong \mathbf{0},$$
$$\mathbf{A} \cdot \mathbf{S} \cdot \mathbf{h} \cong \mathbf{0}, \tag{39}$$

where $\tilde{\mathbf{H}}$ is the extended matrix (38), matrix $\mathbf{S}$ is analogous to (31), its size is $(N_x - L) \cdot (N_y - M) \times L \cdot M$, $\mathbf{h} = \mathbf{vec}(H)$. The distinction of the matrices $\tilde{\mathbf{H}}$ and $\mathbf{S}$ interaction with the operator $\mathbf{A}$ in expressions (39) consists in presence of a window which sequentially extracts one, two and etc. elements of the vector $\mathbf{h}$ up to full vector and then in inverse order. This is equivalent to convolution of the PSF with 2D $\delta$-function window (6). The window tradeoffs the spaces $\Xi$ and $\Omega$ for vector $\mathbf{h}$ and it vanishes if the vector $\mathbf{h}$ is presented in the space $\Xi$. The function $H$ is invariant to the conjugation transform in (39) if it obeys zero bound condition in the space $\Xi$. This feature of the PSF defines restriction on its shape.

The operator $\mathbf{A}$ is given in the space $\Xi \times \Omega$. The sequence order of $\mathbf{h}$ and $\mathbf{A}$ interaction will not change if the vector expressions (34) will be rewritten as a set of scalar values

$$\mathbf{h}^T \cdot \mathbf{A} \cdot \mathbf{s} \cong 0. \tag{40}$$

The expression (40) is the discrete presentation of the second integral expression in (5) and its equivalence to expressions (39) can be inspected by direct calculations. So, the conjugation transform in (5) means the transposition. As it follows from (34), equation (40) is satisfied for arbitrary vector $\mathbf{h}$, but since the vectors $\mathbf{h}$ and $\mathbf{s}$ are convolved components of the vector $\mathbf{x}$ in (36) the vector $\mathbf{h}$ belongs to the left side (conjugated) NS of the operator $\mathbf{A}$:

$$\mathbf{h}^T \cdot \mathbf{A} \cong \mathbf{0}. \tag{41}$$

We have shown above and in subsection 2.1 the sufficient proof of (41) and its analog (9): if the vector $\mathbf{h}$ is the PSF then (41) and (9) are true. Additional conditions for the PSF evaluation were formulated as the spectral filter (25).

It can be notated that the matrix $\mathbf{A}$ in (41) is sufficiently smaller then the same matrix in (34). It includes $L \cdot M$ rows and $(P+L-1)(Q+M-1)$ columns, $L < P$ and $M < Q$. The hypersurfaces $Z_S : S(Z_S) = 0$ and $Z_H : H(Z_H) = 0$ can be separated using a single image fragment because they are mutually independent (incoherent) – $Z_S \cap Z_H = 0$. Their ZT's are given in different spaces – $\Omega$ and $\Xi$. It is known that space and ZT arguments are at the mutually inverse positions. On the other hand, original image $S$ is coherent to measured image $X$ and therefore there were used $K \geq 3$ degraded images with incoherent blur functions for reconstruction of $S$ in [45].

The CNS of the operator (33) can be defined with the help of the singular value decomposition (SVD) of the matrices product $\mathbf{A} \cdot \mathbf{A}^T$ as a set of K orthogonal vectors corresponding to smallest singular values and presented as the matrix $\mathbf{V}_{NS}$ of size $(L \cdot M) \times K$. The product $\mathbf{A} \cdot \mathbf{A}^T$ gives the discrete analog of operator (10). The cross-auto-correlation matrix (18) can be given as the product $\mathbf{R}_{crX} = \nabla \mathbf{X} \cdot (\mathbf{J} \cdot \nabla \mathbf{X})^T$, where matrix $\nabla \mathbf{X}$ is analogous to matrix (31) of the size $(L \cdot M) \times (N_x - L) \cdot (N_y - M)$ and it is compiled by the elements $\nabla \mathbf{x}_{i,k} = 0.5(\mathbf{x}_{i+1,k} - \mathbf{x}_{i-1,k} + \mathbf{x}_{i,k+1} - \mathbf{x}_{i,k-1})$, $\mathbf{J}$ is the cross-diagonal unit matrix. The transform $\mathbf{\rho}_X = \mathbf{V}_{NS}^T \mathbf{R}_{crX} \mathbf{V}_{NS}$ gives the spectral matrix of the size $K \times K$. Same transform of the averaged matrix

$$\nabla \overline{\mathbf{X}} = (N_x - 2L)^{-1}(N_y - 2M)^{-1} \sum_{k=0}^{N_x - 2L - 1} \sum_{n=0}^{N_y - 2M - 1} \left[ \nabla \mathbf{x}_{i+j+k, l+m+n} \right]_{i,j=1...L}^{l,m=1...M}$$

gives the spectral matrix $\overline{\omega}_X$ with the elements analogous to (23). We can evaluate the elements of the vector **v** in (14) basing on (25) as $v_k \cong \gamma \cdot sign(\rho_{X_{k,k}} + \overline{\omega}^2_{X_{k,k}})\sqrt{|\rho_{X_{k,k}} + \overline{\omega}^2_{X_{k,k}}|}$ and the vector **h** elements in (41) as

$$h_{i,k} \cong \sum_{\kappa \in NS} v_\kappa V_{i,k;\kappa} V_{i,k;\kappa}, \qquad (42)$$

where $V_{i,k;\kappa} = V_{i \cdot M + k;\kappa}$ – the elements of the $\kappa - th$ CNS vector, $\gamma$ – the normalization coefficient in accordance with normalization condition (26)

$$\sum_{i=1}^{L} \sum_{k=1}^{M} h_{i,k} = 1. \qquad (43)$$

The discrete manner of IPSF (28) looks as

$$g_{i,k} \cong \sum_{\kappa \in NS} u_\kappa V_{i,k;\kappa} V_{i,k;\kappa}, \qquad (44)$$

where vector **u** is defined with the help of discrete version of equation (30) as the central column of the matrix given by pseudo inversion [40] of the matrices product –

$$\mathbf{u} = (\mathbf{V}_{NS} \mathbf{H})^{\#}_{N/2}, \qquad (45)$$

where $\mathbf{V}_{NS}$ is the matrix of the size $K \times L \cdot M$ compiled by functions $V_{i,k;\kappa} V_{i,k;\kappa} : \kappa \in NS$ samples and **H** is matrix (38) of the size $L \cdot M \times (2L-1)(2M-1)$, $N = (2L-1) \cdot (2M-1)$. The found IPSF can be normalized as well as PSF with the help of condition (43).

### 3. PSF and IPSF optimization

Implementation of deconvolution (2) in the manner of a filter with the finite impulse response is preferable for applications. However, the exact IPSF should be known for such implementation. The inversion transform belongs to ill-posed problems from the point of view of computation and, in addition, is sensitive to small fluctuations of estimated or measured PSF. Therefore we suggest the methods of the PSF and IPSF shape optimization with the aim to eliminate fluctuations caused by approximate feature of their estimation.

The PSF and IPSF estimates (42) and (44) are given with using of image gradient first and second moments without restrictions on the sought blur characteristic shapes. The shapes undesirable fluctuations can be eliminated by optimization of the spectrum vectors in (42) and (44). We suppose that the PSF and IPSF belong to a class of smooth two dimension functions. The functions surface area can serves as a criterion of smoothness because the minimal surface area corresponds to maximally smooth flat surface. The surface area of the function $H(x,y)$ is defined as [19]

$$\Sigma_H = \int_\Xi \sqrt{1 + H_x^2 + H_y^2} (x, y) dx dy, \qquad (46)$$

where $H_x = \partial H(x,y)/\partial x$ and $H_y = \partial H(x,y)/\partial y$. Expression (46) can be considering as regularization functional $\mathbf{R}(H)$ in the general functional of the optimization problem statement.

*3.1. PSF optimization in the CNS spectral domain*
The optimization problem can be presented with the help of the functional which complements equation (15) by restriction on a sought function shape:

$$I_H = \arg\min_{\mathbf{v}} \left\{ \int_\Xi \left\| H(x,y) - \sum_{\kappa \in NS} v_\kappa V_\kappa(x,y) V_\kappa(x,y) \right\|^2 dxdy + \lambda \cdot \mathbf{R}(H) \right\}, \quad (47)$$

where $\lambda$ – the regularization parameter. In accordance with (42) and (46), functional (47) in the discrete manner has the following form.

$$I_h = \arg\min_{\mathbf{v}} \left[ \sum_{i,k=1}^{L,M} \left( \left\| h_{i,k} - \sum_{\kappa \in NS} v_\kappa V_{i,k;\kappa} V_{i,k;\kappa} \right\|^2 + \lambda \sqrt{1 + 4\left(\sum_{\kappa \in NS} v_\kappa V_{x\,i,k;\kappa} V_{i,k;\kappa}\right)^2 + 4\left(\sum_{\kappa \in NS} v_\kappa V_{y\,i,k;\kappa} V_{i,k;\kappa}\right)^2} \right) \right], \quad (48)$$

where $V_{x\,i,k;\kappa}$ and $V_{y\,i,k;\kappa}$ are discrete derivatives of the pointed functions. Euler-Lagrange variational derivative [19] for the functional (48) minimum finding looks as the equations system

$$\frac{\delta I_h}{\delta v_m} \equiv 0: \quad \forall m \in NS. \quad (49)$$

The substitution (48) into expression (49) yields the following equations system for iterative evaluation of the vector $\mathbf{v}$ elements.

$$\sum_{\kappa \in NS} v_\kappa^{(t+1)} \left( \sum_{i,k=1}^{L,M} V_{i,k;\kappa} V_{i,k;\kappa} V_{i,k;m} V_{i,k;m} + 2\lambda \cdot \Phi_{m,\kappa}(\mathbf{v}^{(t)}) \right) = \sum_{i,k=1}^{L,M} h_{i,k}^{(t)} V_{i,k;m} V_{i,k;m}, \quad (50)$$

where $h_{i,k}^{(0)} = h_{i,k}$ in (42), $h_{i,k}^{(t)} \cong \sum_{\kappa \in NS} v_\kappa^{(t)} V_{i,k;\kappa} V_{i,k;\kappa}$,

$$\Phi_{m,\kappa}(\mathbf{v}) = \sum_{i,k=1}^{L,M} \frac{V_{x\,i,k;\kappa} V_{i,k;\kappa} V_{x\,i,k;m} V_{i,k;m} + V_{y\,i,k;\kappa} V_{i,k;\kappa} V_{y\,i,k;m} V_{i,k;m}}{\sqrt{1 + 4\left(\sum_{\kappa \in NS} v_\kappa V_{x\,i,k;\kappa} V_{i,k;\kappa}\right)^2 + 4\left(\sum_{\kappa \in NS} v_\kappa V_{y\,i,k;\kappa} V_{i,k;\kappa}\right)^2}}.$$

The regularization parameter $\lambda$ can be chosen by the help of the condition of convergence of the $q$ first iterations:

$$\sum_{i,k=1}^{L,M} \left\| h_{i,k}^{(t+1)} - h_{i,k}^{(t)} \right\|^2 \theta \leq \sum_{i,k=1}^{L,M} \left\| h_{i,k}^{(t)} - h_{i,k}^{(t-1)} \right\|^2, \quad (51)$$

where $t = 1, 2, .., q$, $\theta$ – a positive value The iteration process can be stopped if

$$\sum_{i,k=1}^{L,M} \left\| h_{i,k}^{(t+1)} - h_{i,k}^{(t)} \right\|^2 \leq \varepsilon_h, \quad (52)$$

where $\varepsilon_h$ is a small value.

*3.2. IPSF optimization in the CNS spectral domain*

The functional of the IPSF optimization problem is basing on the relations between functions $H$ and $G$ (29), (30) and regularization, it is the following.

$$I_G = \arg\min_{\mathbf{u}} \left\{ \int_\Xi \left\| \sum_{\kappa \in NS} u_\kappa (V_\kappa(x,y) V_\kappa(x,y)) * H(x,y) - \delta(x,y) \right\|^2 dxdy + \lambda \cdot \mathbf{R}(G) \right\}. \quad (53)$$

The variational derivative (49) of functional (53) relatively $u_m : \forall m \in NS$ yields with account (45) and (46) the equations system for iterative evaluation of the vector $\mathbf{u}$.

$$\sum_{\kappa \in NS} u_\kappa^{(t+1)} \left( \Psi_{m,\kappa} + 2\lambda \cdot \Phi_{m,\kappa}\left(\mathbf{u}^{(t)}\right) \right) = \sum_{i,k=1}^{L,M} V_{i,k;m} V_{i,k;m} (\mathbf{H})_{i,k;N/2}, \qquad (54)$$

where $\Psi_{m,\kappa} = \sum_{l=1}^{N} \sum_{i,k=1}^{L,M} V_{i,k;\kappa} V_{i,k;\kappa} (\mathbf{H})_{i,k;l} \sum_{i',k'=1}^{L,M} V_{i',k';m} V_{i',k';m} (\mathbf{H})_{i',k';l}$, $(\mathbf{H})_{i,k;l}$ – the elements of matrix (38). The regularization parameter in schema (54) can be chosen as in previous schema (50).

*3.3. IPSF estimation and optimization in image space domain*

We consider the IPSF estimation in image space domain as alternative approach to CNS method. The IPSF estimation is based on Wiener spectral method [5, 30] which includes division by the PSF Fourier spectrum and therefore needs regularization. However, even regularized inverse spectrum gives image boundaries fluctuations. Also, the regularization eliminates filtered image resolution. These deficiencies are caused by discrepancy of the image and PSF spaces $\Omega$ and $\Xi$. The artificial matching of the PSF into $\Omega$ in the spectral transforms by supplementation of zeros creates the spectrum leakage with fluctuations which influence is intensified by division. We shall find the IPSF without exceeding bounds of $\Xi$ to avoid these defects.

Let's write the expressions (1) and (2) in discrete manner.

$$\mathbf{x}_{i,k} = \sum_{l=1}^{L} \sum_{m=1}^{M} h_{l,m} \mathbf{s}_{i+l,k+m}, \qquad (55)$$

$$\mathbf{s}_{i,k} = \sum_{l=1}^{L} \sum_{m=1}^{M} g_{l,m} \mathbf{x}_{i+l,k+m}, \qquad (56)$$

where $\mathbf{x}_{i,k}$ and $\mathbf{s}_{i,k}$ – the samples or RGB vectors of observed distorted image $X$ and sought original image $S$, $h_{l,m}$, $g_{l,m}$ – the elements of the PSF and IPSF, $i = 0...N_x - L - 1$, $k = 0...N_y - M - 1$. The substitution (56) into (55) gives the equation for evaluation of the IPSF.

$$\mathbf{x}_{i,k} = \sum_{l=1}^{L} \sum_{m=1}^{M} g_{l,m} \left( \sum_{l'=1}^{L} \sum_{m'=1}^{M} h_{l',m'} \mathbf{x}_{i+l-l',k+m-m'} \right). \qquad (57)$$

The expression in the parenthesis yields once more degradation of observed image by the filtering. The indices of $\mathbf{x}_\bullet$ in right part of equation (57) are symmetrized relatively to the left part indices due to symmetry of convolution. The equation (57) is ill-posed because $N_x \gg L$ and $N_y \gg M$. The primary solution of the equation (57) can be found by the least squares method.

$$\mathbf{g} = R_{YY}^{\#} \mathbf{r}_{YX}, \qquad (58)$$

where $\mathbf{g} = \mathbf{vec}(G)$: $\mathbf{g} \in \Xi$, $\mathbf{Y} = \mathbf{H} \cdot \mathbf{X}$, $\mathbf{H}$ – the matrix (38), $\mathbf{X}$ and $\mathbf{Y}$ are the extended matrices similar to (31) of the size $(2L-1) \cdot (2M-1) \times (N_x - L - 1) \cdot (N_y - M - 1)$,

$$R_{YY} = \mathbf{Y} \cdot \mathbf{Y}^T : R_{YY} \in \Xi \times \Xi, \quad \mathbf{r}_{YX} = \mathbf{Y} \cdot \mathbf{x} : \mathbf{r}_{YX} \in \Xi. \qquad (59)$$

If the matrix $R_{YY}$ is close to singular one then the solution (58) is unstable and therefore equation (57) needs regularization which applies restrictions on its solution.

We will use functional (46) for regularization in PSF space as well as in spectral domain. The general functional, which includes least squares solution of the equation (57) and regularization term (46), can be written as

$$I_G = \arg\min_G \left[ \frac{1}{2} \int_\Xi \| G * Y - X \|^2 (x,y) dx dy + \lambda \cdot \Sigma_G \right], \qquad (60)$$

where $G_x = \partial G(x,y)/\partial x$ and $G_y = \partial G(x,y)/\partial y$. Euler-Lagrange variational derivative [19] for the functional (60) minimum finding is the following.

$$\frac{\partial \|\mathbf{G} \cdot \mathbf{Y} - \mathbf{X}\|^2}{2\partial G} - \lambda \cdot \left( \frac{\partial}{\partial x} \frac{\partial}{\partial G_x} + \frac{\partial}{\partial y} \frac{\partial}{\partial G_y} \right) \sqrt{1 + G_x^2 + G_y^2} = 0, \quad (61)$$

where $\mathbf{G}$ is the matrix similar to (38). We also mean fulfillment of zero Neumann boundary condition [19]. With the account of (61), the expression (60) takes the next form.

$$(\mathbf{G} \cdot \mathbf{Y} - \mathbf{X}) \cdot \mathbf{Y}^T - \lambda \cdot \mathbf{L}_\Sigma(G) = 0, \quad (62)$$

where

$$\mathbf{L}_\Sigma(G) = \left(1 + G_x^2 + G_y^2\right)^{-3/2} \left(\left(1 + G_y^2\right) G_{xx} + \left(1 + G_x^2\right) G_{yy} - 2 G_x G_y G_{xy}\right). \quad (63)$$

The discrete iterative schema of the equation (62) solution is the following:

$$\mathbf{g}^{(k+1)} = \left(R_{YY} - \lambda \cdot \delta R\left(\mathbf{g}^{(k)}\right)\right)^{-1} \mathbf{r}_{YX}, \quad (64)$$

$$\delta R\left(\mathbf{g}^{(k)}\right) = \left(\mathbf{I} + diag\left[\left(D_x \mathbf{g}^{(k)}\right)^2 + \left(D_y \mathbf{g}^{(k)}\right)^2\right]\right)^{-3/2} \times$$

$$\left(\left(\mathbf{I} + diag\left[\left(D_y \mathbf{g}^{(k)}\right)^2\right]\right) D_{xx} + \left(\mathbf{I} + diag\left[\left(D_x \mathbf{g}^{(k)}\right)^2\right]\right) D_{yy} - 2 \cdot diag\left[D_x \mathbf{g}^{(k)}\right] \cdot diag\left[D_y \mathbf{g}^{(k)}\right] D_{xy}\right),$$

$\mathbf{g}^{(0)} = \mathbf{g}$ in (58), $\mathbf{I}$ – the identity matrix, $D_\bullet$ – the differential operator with pointed by indices variables and order, $diag[\cdot]$ – the diagonal matrix, created by the vector in brackets. The parameter $\lambda$ can be chosen as maximal possible value that provides convergence of the schema (64) at first $q$ iterations, as it was shown in (51). The iteration process can be stopped on condition similar to (52).

The IPSF (64) can be used for primary estimation of sought original image with the help of discrete convolution (56). Of course, the spectral method can be also used but in this case the division by the IPSF spectrum is not necessary and influence of spectrum leakage is insignificant.

It is known that the surface area as regularization functional was used in image denoising problem solution [3, 50]. The problem (60) and similar denoising and deconvolution problems considered in [3, 50] differ by statement of the problem. Arguments and optimization parameter of the second ones belong to an image space. The arguments and optimization parameter of the functional (60) are defined in different spaces – the arguments $X, Y \in \Omega$ and the optimization parameter $G \in \Xi$. We project $\Omega \to \Xi$ by convolution operations (59) and in that way create the unified problem statement space in $\Xi$.

## 4. Deconvolution optimization

It is difficult to define optimal image estimate by single convolution transform even the IPSF is given exactly. Therefore we consider two approaches to implementation of the fast iterative schemas of image estimate optimization using the maximum entropy (ME) generalization of the least squares approximation of degraded image and regularization technique.

### 4.1. The problem of deconvolution optimization

Original image estimate should complies requirement of optimality in the manner of the functional

$$I = \arg\min_{S,H} \frac{1}{2} \int_\Omega \|X - H * S\|^2 (x, y) dx dy. \quad (65)$$

Process of the functional (65) optimization is usually iterative and it depends on statistical properties of given initial image and sought original one, especially, their conditional probability distribution (PD). The iterative LR algorithm [38, 48] with procedure of the PSF estimate optimization [24] gives a solution of the deconvolution problem in accordance with criterion of the ML of observed $X$ and sought original images $S$. LR algorithm is basing on a maximization of Poisson conditional PD of blurred and original images using Bayes equation. Since the algorithm has slowly convergence the methods of its acceleration there were offered in [7, 47, 52]. As it was shown in [17], LR algorithm does not support convergence if an image PD is not Poisson. Especially in the case of Gaussian PD

$$P(X \mid S) \propto \exp(-\|X - H * S\|^2 / 2\sigma_X^2), \tag{66}$$

where $\sigma_X$ is the image deviation. The ME generalization of the conditional PD (66) gives the functional (65) [17] which is trivial from the point of view of the optimization. Therefore an additional term is introducing usually into (65). The term constitutes some prior image feature. Mostly often is using the total variation regularization functional with various norm of the gradient

$$\mathbf{R}_{TV}(S) = \int_\Omega \|\nabla S\|^\alpha dxdy, \tag{67}$$

where $\alpha = 0.5;\ 0.8;\ 1;\ 2$ [13, 15, 36, 49]. We consider the using of sought image geometric prior in the manner of minimal surface as additional term in (65) or as intrinsic characteristic of the image space. The aim is to design quickly convergent iterative schemas of the deconvolution optimization which are not sensitive to estimated PSF and IPSF fluctuations.

*4.2. The method of deconvolution optimization by balanced variations and dynamic regularization (BVDR)*

The deconvolution optimization (65) with additional regularization term $\mathbf{R}(S)$ means the minimization of the functional

$$I_S = \arg\min_S \left[ \frac{1}{2} \int_\Omega \left( \|X - H * S\|^2 + \lambda \cdot \mathbf{R}(S) \right)(x, y) dxdy \right], \tag{68}$$

where $\lambda$ – the regularization parameter, a positive value or function. The EL equation gives for the functional (68) the expression

$$H'*(X - H * S) + \lambda \cdot \mathbf{L}(S) = 0, \tag{69}$$

where $\mathbf{L}(S) = \delta \mathbf{R}(S)/\delta S$ – the variational derivative that is analogous to the second term in (61). We suppose that the convolution $G * H'$ gives the trivial operator. Then the convolution of the IPSF with components of (69) gives the next equation:

$$X - H * S + \lambda \cdot G * \mathbf{L}(S) = 0. \tag{70}$$

We can parameterize the equation (70) and present it as evolutional equation in accordance with gradient-descent method [32, 49],

$$S_t(t) = X - H * S(t) + \lambda(t) \cdot G * \mathbf{L}(S(t)), \tag{71}$$

where $t$ – the evolution parameter. Neumann boundary conditions are assumed. The regularization parameter $\lambda(t)$ can be defined as such that equation (71) has to be convergent.

The convergence of iterative problems in image processing was considered in review paper [27] and some other papers [4, 11, 22, 25, 34]. We can define convergence process with the help of $l_1$ norm as $\langle |S(t+\tau) - S(t)| \rangle_\Omega \leq \langle |S(t) - S(t-\tau)| \rangle_\Omega : \tau \to 0$, or as parametric derivative

$$\langle |S_t(t)|_t \rangle_\Omega \leq -\chi, \tag{72}$$

where $\chi$ is a positive small value, $\langle \cdot \rangle_\Omega$ – the averaging in the image space $\Omega$. We can write expression (72) for (71) as the following:

$$\langle |-H * S(t) + \lambda(t) \cdot G * \mathbf{L}(S(t))|_t \rangle_\Omega \geq \chi. \tag{73}$$

Since we have found the regularization parameter as a positive function the equation (73) can be rewritten as

$$\lambda_t(t) \cdot \langle |G * \mathbf{L}(S(t))| \rangle_\Omega + \lambda(t) \cdot \langle |G * \mathbf{L}(S(t))|_t \rangle_\Omega + \langle |-H * S(t)|_t \rangle_\Omega \geq \chi. \tag{74}$$

The $\chi$ is arbitrary value and so the expression (74) can be transformed into the differential equation

$$\lambda_t(t) + \lambda(t) \frac{\langle |G * \mathbf{L}(S(t))|_t \rangle_\Omega}{\langle |G * \mathbf{L}(S(t))| \rangle_\Omega} = \frac{\langle |H * S(t)|_t \rangle_\Omega - \chi}{\langle |G * \mathbf{L}(S(t))| \rangle_\Omega},$$

which solution yields the upper bound of $\lambda(t)$:

$$\sup(\lambda(t)) \leq \lim_{\chi \to 0} \int_0^t \exp\left(-\int_\tau^t \frac{\langle |G * \mathbf{L}(S(\varsigma))|_\varsigma \rangle_\Omega}{\langle |G * \mathbf{L}(S(\varsigma))| \rangle_\Omega}(\varsigma) d\varsigma\right) \frac{\langle |H * S(\tau)|_\tau \rangle_\Omega - \chi}{\langle |G * \mathbf{L}(S(\tau))| \rangle_\Omega}(\tau) d\tau +$$

$$\lambda(0) \exp\left(-\int_0^t \frac{\langle |G * \mathbf{L}(S(\tau))|_\tau \rangle_\Omega}{\langle |G * \mathbf{L}(S(\tau))| \rangle_\Omega}(\tau) d\tau\right). \tag{75}$$

The convergence condition (72) in discrete manner means that $\langle |S^{(k+1)} - S^{(k)}| \rangle_\Omega < \langle |S^{(k)} - S^{(k-1)}| \rangle_\Omega$. This condition provides the fulfillment of convergence condition with $l_2$ norm of the residuals like (51)

$$\langle \|S^{(k+1)} - S^{(k)}\|^2 \rangle_\Omega \theta \leq \langle \|S^{(k)} - S^{(k-1)}\|^2 \rangle_\Omega \tag{76}$$

because image is a positive function.

The formula (75) and the equation (71) in discrete form are the following:

$$\lambda^{(0)} = \frac{\langle |H * (S^{(0)} - X)| \rangle_\Omega}{\delta t \langle |G * \mathbf{L}(S^{(0)})| \rangle_\Omega} \left( \exp\left(\frac{\langle G * \left(|\mathbf{L}(S^{(0)})| - |\mathbf{L}(X)|\right) \rangle_\Omega}{\delta t \langle |G * \mathbf{L}(S^{(0)})| \rangle_\Omega}\right) - 1 \right)^{-1}; \tag{77}$$

$$\lambda^{(k)} = \left( \lambda^{(k-1)} + \frac{\left\langle \left| H * \left( S^{(k)} - S^{(k-1)} \right) \right| \right\rangle_\Omega}{\delta t \left\langle \left| G * \mathbf{L}(S^{(k)}) \right| \right\rangle_\Omega} \right) \exp\left( - \frac{\left\langle G * \left( \left| \mathbf{L}(S^{(k)}) \right| - \left| \mathbf{L}(S^{(k-1)}) \right| \right) \right\rangle_\Omega}{\delta t \left\langle \left| G * \mathbf{L}(S^{(k)}) \right| \right\rangle_\Omega} \right),$$

$$S^{(k+1)} = S^{(k)} + \delta t \cdot \left( X - H * S^{(k)} + \lambda^{(k)} \cdot G * \mathbf{L}(S^{(k)}) \right), \tag{78}$$

where $S^{(-1)} = X$, $S^{(0)} = G * X$ and $\delta t$ – the relaxation parameter.

The iterative schema (78), in contrast to known ones, includes convolutions with the PSF and IPSF which are competitive processes of smoothing and peaking of image shape. If we assume that $\lambda_t(t >> 0) \approx 0$ in (74), then approximate estimate of the regularization parameter will be the following:

$$\lambda^{(k)} \approx \frac{\left\langle \left| H * (S^{(k)} - S^{(k-1)}) \right| \right\rangle_\Omega}{\left\langle G * \left( \left| \mathbf{L}(S^{(k)}) \right| - \left| \mathbf{L}(S^{(k-1)}) \right| \right) \right\rangle_\Omega}. \tag{79}$$

The expression (79) shows that the equation (70) constitutes a balance of variations which are caused by convolutions of the PSF and IPSF with sought image estimate and the regularization term. The first one smooth variation of image surface and the second one sharpen contours since the regularization operator is a function of image derivatives. These processes mutually weaken an influence of the PSF and IPSF estimates fluctuations.

The schema (77)-(78) has dynamic character because the regularization parameter depends on current and previous variations of the image estimate: $\lambda^{(k)}$ is the function of $\lambda^{(k-1)}$.

*4.3. The method of deconvolution optimization in curved space (CS)*

The target of optimization of the functional like (65) is to define such $S_{opt}$ that [19]

$$\frac{\delta I(S_{opt})}{\delta S} \equiv 0. \tag{80}$$

In other words, the optimal value of the functional is an invariant of argument fluctuation. Let's consider the functional (65) in discrete form for $S = S_{opt}$.

$$\lim_{\Delta S_{opt} \to 0} \frac{1}{2} \sum_{i,k} \left\| X - H * S_{opt} \right\|^2 \mathrm{P}(\Delta S_{opt})(x_i, y_k) = \min, \tag{81}$$

where $\mathrm{P}(\Delta S_{opt}) = \Delta x \Delta y$ – the projection of image surface element on the plane $XOY$. The variation of expression (81) with account of (80) yields the term

$$\lim_{\Delta S_{opt} \to 0} \frac{1}{2} \sum_{i,k} \left\| X - H * S_{opt} \right\|^2 \frac{\partial \mathrm{P}(\Delta S_{opt})}{\partial \delta S}(x_i, y_k) \tag{82}$$

which may belongs to a zero neighborhood as a balanced sum whereas it can be nonzero in image points. If $\Delta S_{opt}$ will be included fully, not by its projection, then this term will be identical to zero in accordance with condition (80). The surface element in CS with induced by the surface metric is an invariant of variational transforms which cause corresponding tensor transforms of coordinates. It has the following manner [19]: $\Delta S_{opt} = \sqrt{\sigma(S_{opt})} \Delta x \Delta y$, where $\sigma(S) = 1 + S_x^2 + S_y^2$ – the determinant of the metric tensor in a point of $\Delta S_{opt}$.

The problem of functional (65) optimization with the account of space curvature can be written as

$$I_S = \arg\min_S \left[ \frac{1}{2} \int_\Omega \|X - H * S\|^2 \sqrt{\sigma(S)}(x, y) dx dy \right]. \tag{83}$$

The EL equation gives for (83) the next expression:

$$H'*(X - H*S) - \frac{1}{4\sigma(S)}(X - H*S)^2 \left( \frac{\partial}{\partial x}\frac{\partial \sigma(S)}{\partial S_x} + \frac{\partial}{\partial y}\frac{\partial \sigma(S)}{\partial S_y} \right) = 0. \tag{84}$$

The convolution of (84) with the IPSF and its parameterization give the iterative schema for the problem (83) solution.

$$S^{(k+1)} = S^{(k)} + \delta t \cdot \left( X - H*S^{(k)} + G * \left( \Lambda(H, S^{(k)}) \cdot \mathbf{L}_\Sigma(S^{(k)}) \right) \right), \tag{85}$$

where

$$\Lambda(H, S^{(k)}) = \frac{1}{2\sigma(S^{(k)})} \left( X - H*S^{(k)} \right)^2, \tag{86}$$

the operator $\mathbf{L}_\Sigma(S^{(k)})$ is similar to (63). As it follows from the expressions (85) and (86), the iterative process in every point of image surface is assigned by its own parameter – the corresponding element of the regularization surface $\Lambda$. So, we have obtained the schema of total optimization – optimization in each image point, in contrast to schemas of an integral values optimization like (58) and (78).

Let's consider the convergence feature of the schema (85) with using the condition (72). It can be written for (85) as

$$\left\langle \left|S_t(t)\right|_t \right\rangle_\Omega = \left\langle \left|H*S(t)\right|_t \right\rangle_\Omega + \left\langle \left|G * \left( \Lambda(H, S(t)) \cdot \mathbf{L}_\Sigma(S(t)) \right) \right|_t \right\rangle_\Omega \leq -\chi. \tag{87}$$

We can define the upper bound of the regularization parameter surface with the help of convolution (87) with $H'$ and under the assumption that $\Lambda(H, S(t)) \geq 0$ and $\left|\Lambda(H, S(t>>0))\right|_t \approx 0$ as

$$\sup(\Lambda(H, S(t))) \leq \lim_{\chi \to 0} \frac{\left\langle H'*H * |S(t)|_t - \chi \right\rangle_\Omega}{\left\langle \left|\mathbf{L}_\Sigma(S(t))\right|_t \right\rangle_\Omega}. \tag{88}$$

The substitution of (86) into (88) gives the follows inequality in discrete form.

$$\left\langle \left(X - H*S^{(k)}\right)^2 \right\rangle_\Omega \leq 2\left\langle \sigma(S^{(k)}) \right\rangle_\Omega \frac{\left\langle H'*H * (S^{(k)} - S^{(k-1)}) \right\rangle_\Omega}{\left\langle \left|\mathbf{L}_\Sigma(S^{(k)})\right| - \left|\mathbf{L}_\Sigma(S^{(k-1)})\right| \right\rangle_\Omega}. \tag{89}$$

The asymptotic values of the differences in the numerator and denominator of the fraction in (89) tend to zero and therefore the fraction value tends to one because it is usually assumed that $H$ is normalized: $\left\langle H'*H \right\rangle_\Xi \approx 1$. The left side sentence characterizes a blur level. So, the convergence will be met if the blur level is bounded by double averaged value of image element surface area square. If blur causes image surface variation $X \sim S + \delta S$, where $\|S\|^2 > \|\delta S\|^2$, then, as it follows from (46), the condition (89) occurs always and schema (85) is quite convergent in each image surface point.

We can rewrite the expression (85) relative to $\Delta X^{(k)} = X - H*S^{(k)}$ as

$$\left(\Delta X^{(k)}\right)^2 + \frac{2\sigma\left(S^{(k)}\right)}{\mathbf{L}_\Sigma\left(S^{(k)}\right)}\left(\Delta X_H^{(k)} - \frac{\Delta S_H^{(k)}}{\delta t}\right) = 0, \tag{90}$$

where $\Delta X_H^{(k)} = H'*\Delta X^{(k)}$, $\Delta S_H^{(k)} = H'*\Delta S^{(k)}$, $\Delta S^{(k)} = S^{(k+1)} - S^{(k)}$. Let $\Delta X_H^{(k)} \approx \Delta X^{(k)}$ and $\Delta S_H^{(k)} \approx \Delta S^{(k)}$, then the expression (90) is the quadratic equation and local minimum of the functional (83) will be met in a neighborhood of its real roots, or, approximately, in vicinity of

$$S^{(opt)} \approx G * \left( X \pm \sqrt{\frac{2\sigma\left(S^{(opt)}\right)}{\mathbf{L}_\Sigma\left(S^{(opt)}\right)} \frac{\Delta S^{(opt)}}{\delta t}} \right), \tag{91}$$

As it follows from (91), the convergence of schema (85) may be limited. The condition (76) violates when a root $\Delta X^{(opt)}$ of equation (90) is traversed – it was reached the condition $\left\langle \left\|S^{(k+1)} - S^{(k)}\right\|^2 \right\rangle_\Omega \leq \left(\Delta X^{(opt)}\right)^2$. The expression (91) shows that the optimal solution is equal to the primary estimate with variation in accordance to the regularization operator.

The precision of evaluation of the optimal estimate $S_{opt} \cong S^{(opt)}$ in (80) is conditioned by the relaxation parameter in (91), also in (85). The choosing of $\delta t$ has to be made with the account that the function $\mathbf{L}_\Sigma(S^{(k)})$ values can belong to zero neighborhood. So, the point of $S^{(opt)}$ will be stable on condition that corresponding values of $\Delta S^{(k)}/\delta t$ belong to zero neighborhood too. This condition gives the lower bound of relaxation parameter region of allowable values:

$$\delta t \geq \frac{\left\langle \left|S^{(k+1)} - S^{(k)}\right| \right\rangle_\Omega}{\left\langle \left|\mathbf{L}_\Sigma(S^{(k)})\right| \right\rangle_\Omega}. \tag{92}$$

*Remark*. The functional (83) cannot be used for optimization of the IPSF instead (60) because in this case the condition of unified problem statement space is violated, the regularization surface (86) $\Lambda(G,S) \in \Omega$ and operator (63) $\mathbf{L}_\Sigma(G) \in \Xi$ concurrently.

## 5. Methods implementation and test examples

The results of numerical experiments which confirm the offered in the sections 2.1 and 2.3 approach to the PSF estimation are presented. There were used test images with known blur feature. Also, there were used some images of different structure degraded by different types of blur, their originals are unknown. This is the natural situation in a video basing measuring process. We have investigated significance of the procedures (50), (54) and (64) of the PSF and IPSF optimization, the role of balanced variations in schema (78) and the fast achievement of a local minimum by the schema (85). Also, the deblurring of noised image was considered. The given results have been compared with the same results of the iterative method A. Levin at all [36] of the PSF estimation and image reconstruction which is the best for some types of blur and images that accept sparsity presentation. The blind assessment of the deblurred images quality technique based on image anisotropy analysis was used for comparing of the obtained results. The method of anisotropy index (AI) evaluation was suggested in [21]. The method is basing on spectral analysis of one-dimension Wigner distribution along some directions. We used data vectors of eight points size in the fore directions: $0°$; $45°$; $90°$; $135°$, for each pixel of fragment of size $100 \times 100$ in the processing image centre for evaluation of the AI.

### 5.1. PSF estimation

The first step of PSF estimation is the evaluation of the vector **a** using the equations system (37) with the vector in the right part, which corresponds to the vector **a** central element that is assumed as equal to one. The image presentable part of the size not least $2 \cdot P \cdot Q \times 2 \cdot P \cdot Q$ may be used for evaluation of the vector **a** elements if blur is uniform. At the next step matrix **A** (23) of the size $L \cdot M \times (P + L - 1) \cdot (Q + M - 1)$ has been compiled. The numerical experiments have shown that because the matrix **A** has especial structure the singular values distribution $\lambda_i > \lambda_{i+1}$ of the product $\mathbf{A} \cdot \mathbf{A}^T$ has two parts where $\lambda_i \ll 2\lambda_{i+1}$ and $\lambda_i \gg 2\lambda_{i+1}$. The bound between the parts can be chosen as the bound of the model operator eigen and null spaces. Approximately the bound may be defined as

$$\lambda_{bound} \leq 0.5\lambda_1 \tag{93}$$

At the third step the CNS has to be found by the SVD and the PSF (42) can to be valued and optimized by procedure (50).

The choosing of the AR model (37) order $P \times Q$ and the PSF $H$ size $L \times M$ is primary heuristic. The PSF size must be agreed with the blur level. The model order is chosen correctly if the estimated PSF has bound in zero neighborhood and strongly convex one or more maximums. If the image degradation level is not known the correlation method [10] of the AR model order estimation can be used. The method assumes that the maximal position number of a local peak in the last column of the inverse correlation matrix along axis points on the AR model optimal order value for this coordinate. The PSF size should be chosen 2-3 times lower than the AR model order.

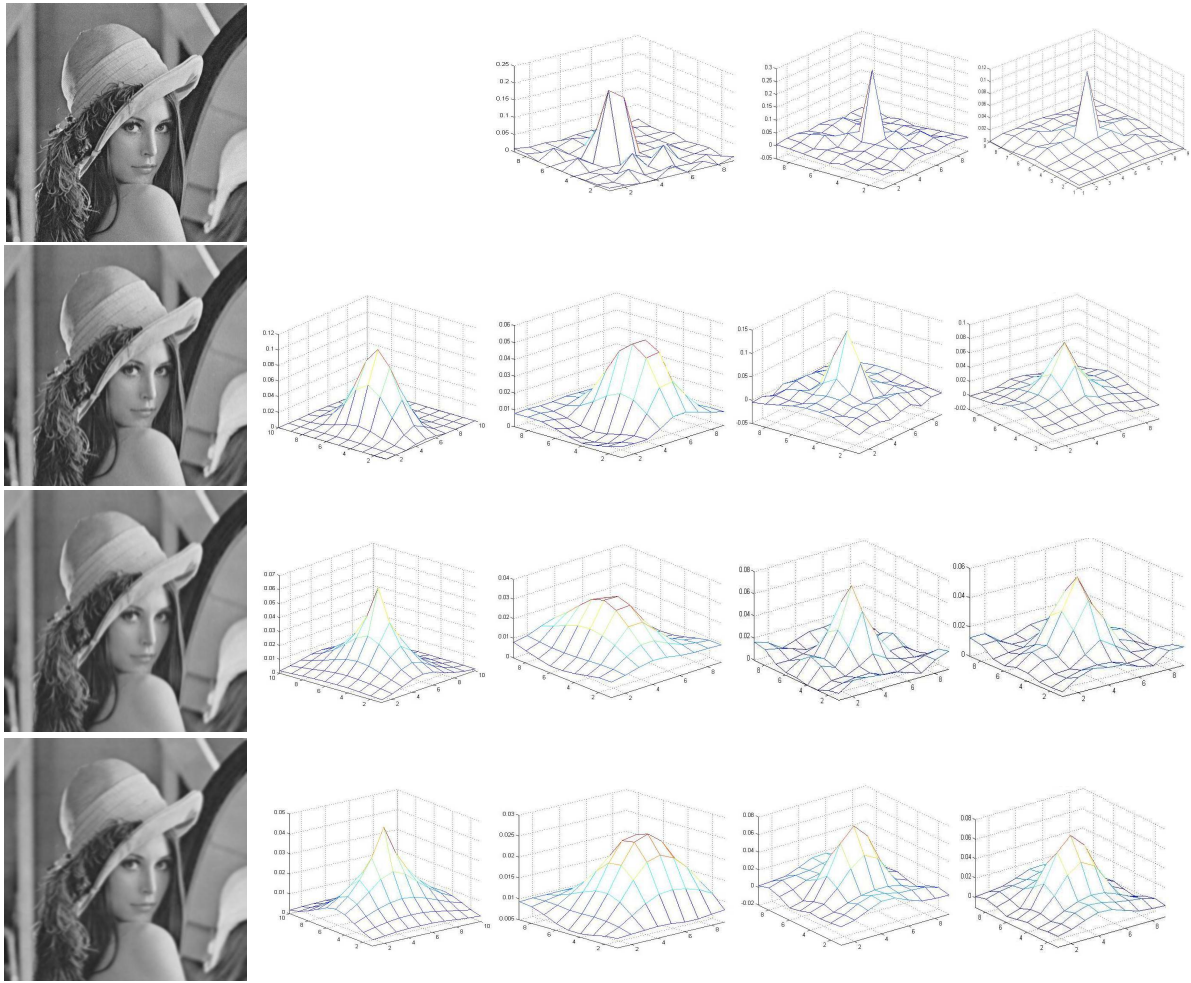

**Figure 1.** Original image at the left in the first row, its AI=0.166839, is blurred by the PSF in the second column. AI's of the blurred images: 0.158151; 0.152816; 0.150236. The third, fourth and fifth columns present the PSF estimates by the method [36] and by the CNS method (41)-(42), PSF with optimized shape by schema (50).

The test image of size $512 \times 512$ pixels is presented in figure 1 high row. The estimates of the PSF given by the method [36] and the CNS method (41)-(42) are presented too. The method [36] is iterative, 22 iterations there where used. The AR model (37) order and the PSF size are the following: $P \times Q = 17 \times 17$; $L \times M = 9 \times 9$. The CNS was bounded by the condition (93) and it was created by 76 eigenvectors of the operator matrix $\mathbf{A}$. The sought PSF is close to trivial – $\delta$-function. The given by the CNS method PSF method is more close to original one. The PSF shape optimization (50) allows to smooth fluctuations which are present in the previous PSF estimate (42). The process of optimization (50) in all experiments was fast ($\theta > 10$), the condition (52) where $\varepsilon_h \leq 10^{-8}$ has to be reached by not more 10 steps at $\lambda = 0.01$.

The rows below show the test image corrupted by blur defined by the PSF which are presented in the second column. In the third-fifth columns are presented PSF estimates. The PSF's given by the method [36] have smoothed central peak and more steep sides along of the PSF shape elongation. The estimated by the CNS method PSF is more close to real PSF in the centre part which is mostly significant. Their wide and height correspond to blur level which is characterizing by the AI. This feature is given by the CNS dimension defined with the help of (93) – 15, 10 and 6 eigenvectors for PSF's in the second, third and fourth rows correspondingly.

### 4.2. IPSF estimation

The estimation of a PSF is only the first step of an image reconstruction. The problem consists in an IPSF estimation and deconvolution (2) in large. It is mostly important to obtain an IPSF adequate to real image spreading nature. So, we will consider the role of an IPSF estimation and optimization by using of the expressions (44), (54) and (58), (64).

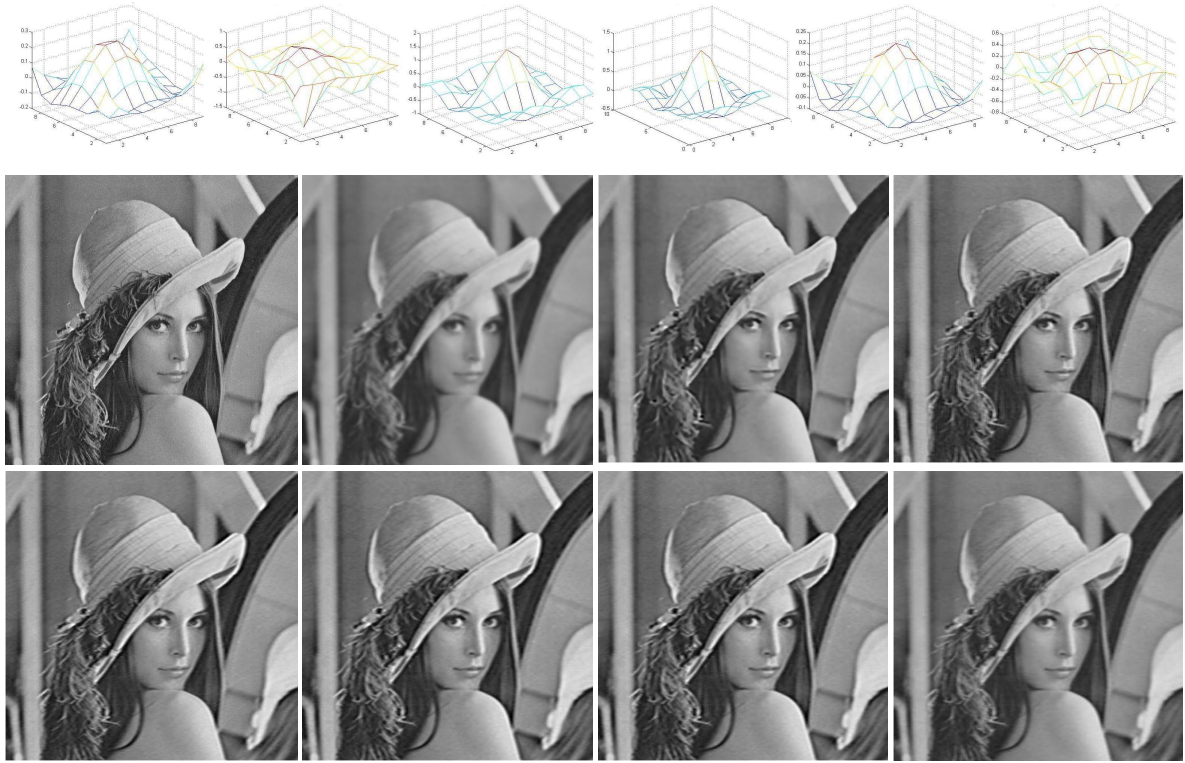

**Figure 2.** From the high row to low and from the left to right are the IPSF's, optimized IPSF's and image estimates given for the image in figure 1 fourth row: IPSF (58) and optimized IPSF (64) given by inversion of the method [36] PSF; two IPSF's (44) with spectral vectors (45) and (54); IPSF (58) and optimized IPSF (64) given by inversion of the CNS method PSF; second row – original and blurred images, reconstructed images by the method [36] (AI=0.167077) and by convolution with optimized IPSF (second in the high row, AI=0.165864); third row – images given by convolution with fore last IPSF's in the high row, AI's are the following: 0.226463; 0.196361; 0.164193; 0.165718.

Figure 2 shows IPSF's which are found for the estimated PSF's in figure 1 lowest row. The images given by the reconstruction of the image in the same row by the method [36] and by convolution (56) with using of the IPSF's are presented too. As it seen from the figure, the optimized IPSF's by space schema (64) have more sharpened form in comparison with non-optimized ones. This feature action appears as higher anisotropy indices of the reconstructed images. As the result, the anisotropy of the reconstructed by the iterative method [36] image has the same level as images that were reconstructed by the single convolution with optimized IPSF. The presented results show that the reconstruction does not strongly depend from the estimated PSF shape if the corresponding IPSF is optimized in accordance with criterion of minimal surface (44). The iterative process of the IPSF's optimization was executed by up to 7 steps with the parameters in (64) and (51), (52): $\lambda \leq 0.01$; $q = 3$; $\theta \leq 10$; $\varepsilon_h = 1.e-8$. The optimization of the CNS spectrum (54) yields smoothed IPSF shape and corresponding image estimate.

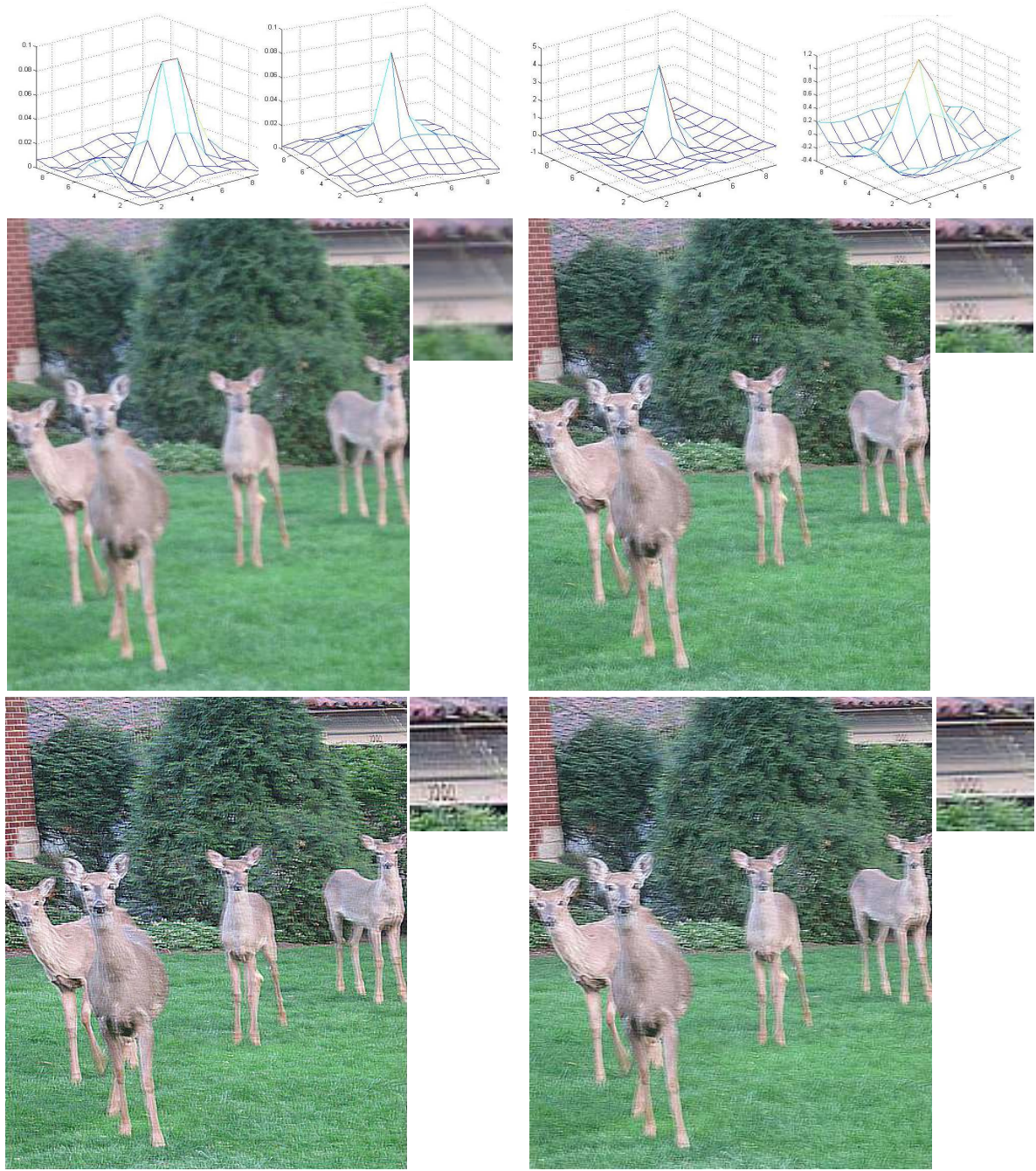

**Figure 3.** Initial image and its zoomed fragment in the right high corner (AI=0.125147) are presented in the first column of the second row. The method [36] PSF, the CNS optimized method PSF and IPSF given by the CNS method with optimization procedure (54), IPSF (58) are presented in the first row (from the left to right). The reconstructed images and zoomed fragments with AI's 0.242511, 0.446511 and 0.476770 given by the method [36] and convolution with two IPSF's presented in the first row are shown in the second and lower rows.

The image in figure 3 includes different natural textures. In this case the process of the PSF's optimization (64) was not successful because the appropriate regularization parameter $0.01 \geq \lambda \geq 1.e-5$ for the condition (51) fulfilling was not found. The AR model order and the PSF size are the same as in the previous example. The CNS of the model operator was created by 47 eigenvectors. The estimated by the method [36] and CNS method PSF's have same elongation and point on blur caused by horizontal moving. The AI's of the reconstructed images show that in the case when image has not sparsity presentation, which can be given with the help of filters like differential ones [36], the PSF estimate by the CNS method is more exact. The images anisotropy can be assessed visually with the help of the small textures appearance in the high right corner.

The initial image in figure 4 is blurred by defocusing. The estimated by the CNS method PSF with optimization (50) has shape close to symmetrical. It is created by 8 eigenvectors of the CNS. The optimized IPSF's are close to symmetrical too. The parameters of the iterative processes (50), (54) and (64) are same as in the examples above.

The examples in figures 3 and 4 have shown that the IPSF's given by the space method (58), (64) are more sharpened in comparison with IPSF (44), correspondingly, the image estimates are more sharpened too. But the central bump wide and shape of the IPSF's given by the spectral CNS method is more close to PSF's bump. Therefore we may suppose that this method is more exact then the space method. Also it can be noted that IPSF estimate can be found using discrete form of equation (29). Such approach does not give appropriate solution. These results confirm PSF and IPSF belonging to the CNS. The space method of IPSF evaluation is preferable in cases when CNS space dimension $K \leq 3$.

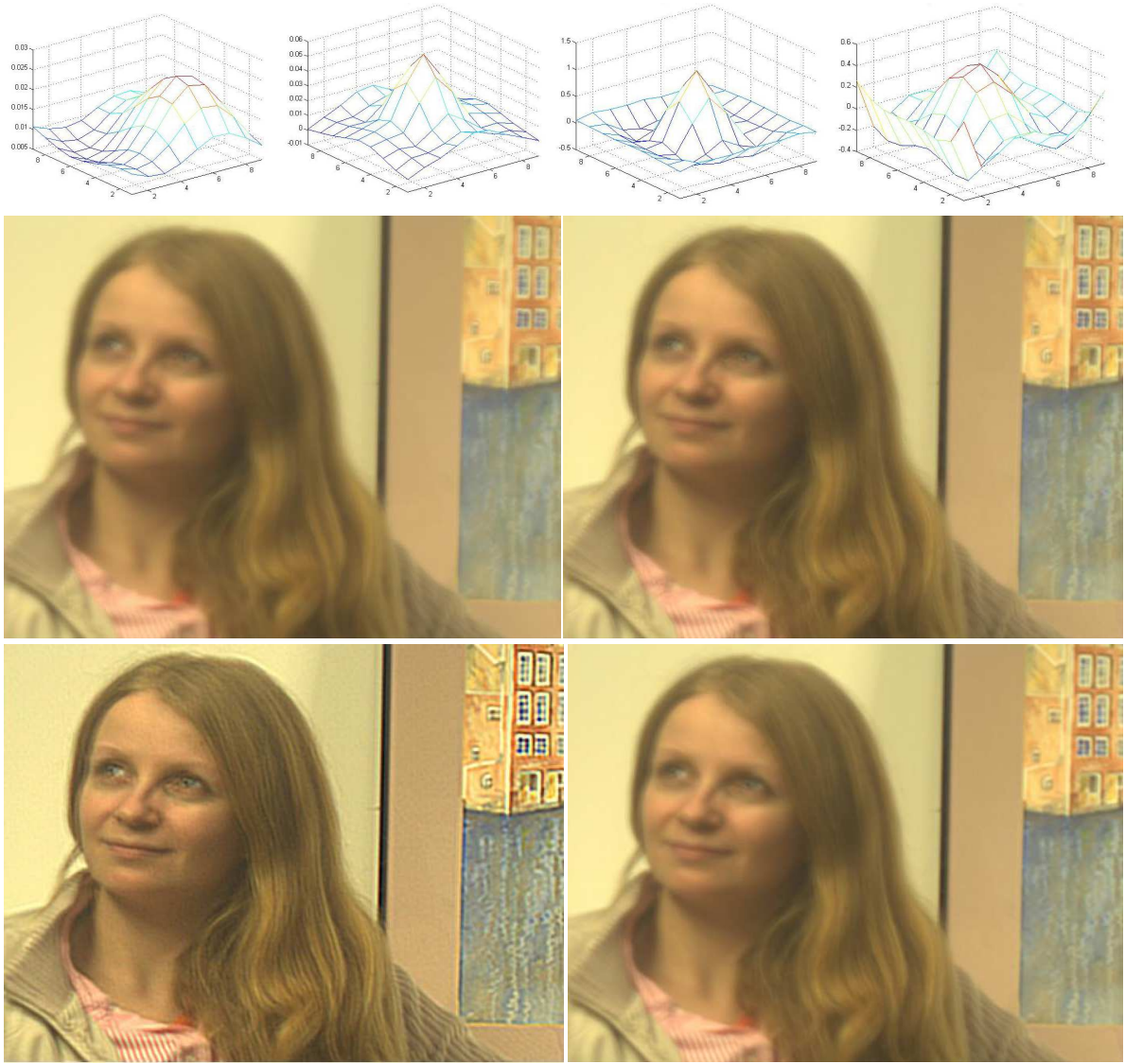

**Figure 4.** Initial image with AI=0.010579 is presented in the first column of the second row. The method [36] PSF; the CNS method optimized PSF and IPSF given by the CNS method with optimization procedure (54), IPSF (64) are presented in the first row (from the left to right). The reconstructed image by the method [36] with AI=0.023659 is shown in the second column of second row. The images with AI's 0.052318 and 0.019625 given by convolution with IPSF's presented in the first row are shown in the third row.

## 4.3 The examples of deconvolution optimization

Deconvolution optimization is intended for finding an image estimate which is maximally close to initial one and satisfies given bounding conditions. The optimization process should compensate inaccuracies in PSF and IPSF estimation, convolution implementation as limited discrete sum. We investigate the iterative schemas (78) and (85) using examples of images in figures 1, 3 and 4.

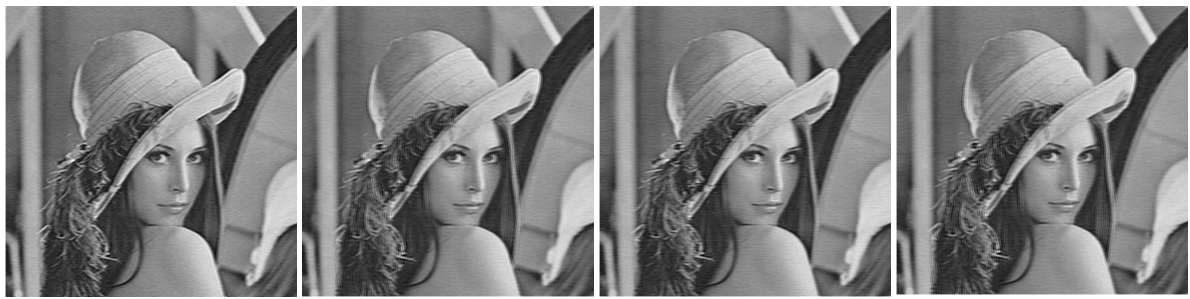

**Figure 5.** Images given by the image in figure 2 second column of the third row optimization using BVDR and CS methods with using of optimized IPSF (44) – two first images with AI's 0.179130, 0.165362, and IPSF (64) – second two images with AI's 0.171267, 0.165211.

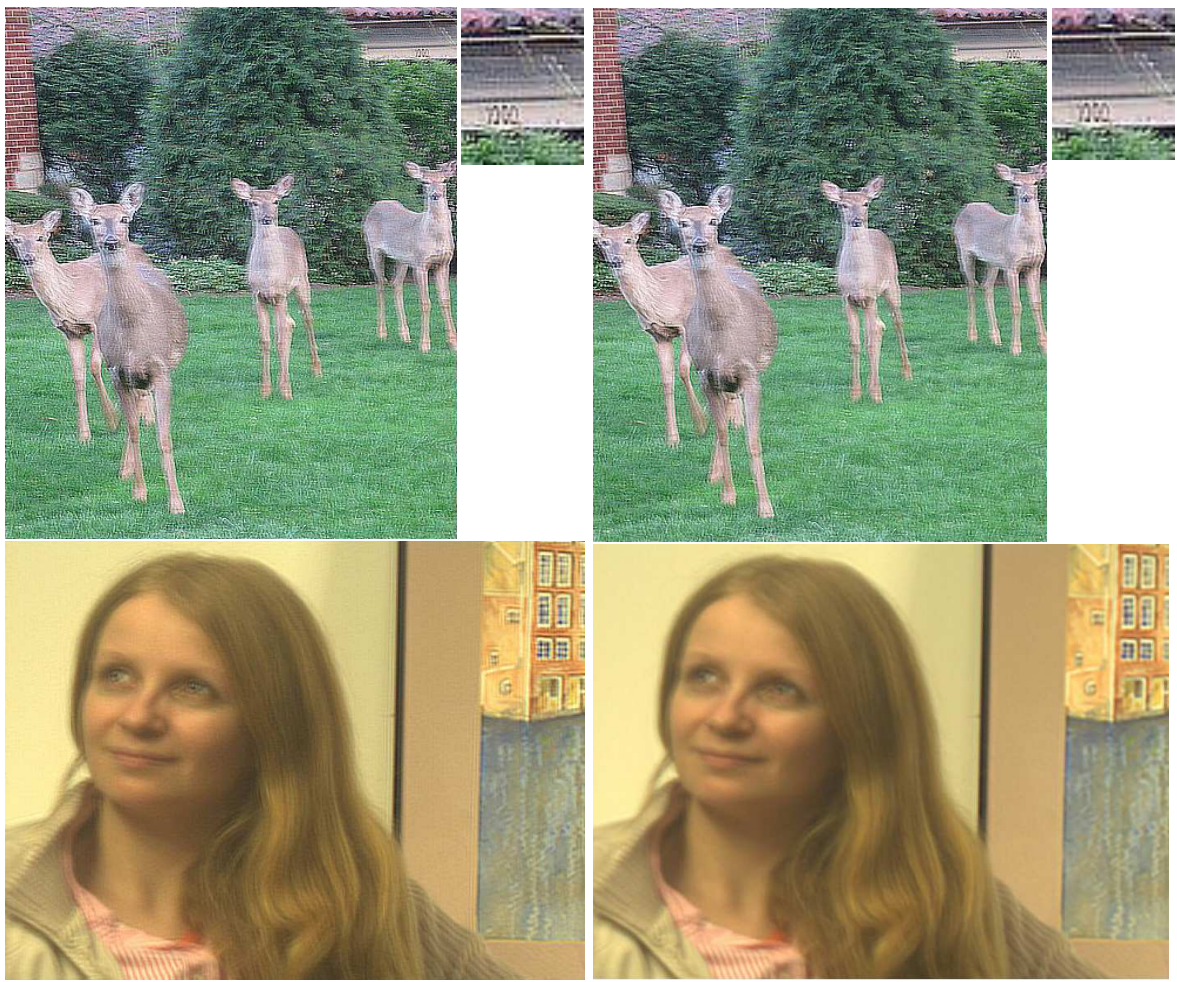

**Figure 6.** Optimized images in figures 3 and 4 first column of third row are defined by the BVDR method (left) and CS method (right). Their AI's are the next: 0.444337, 0.405770 and 0.036605, 0.023655. There were used 20 iteration steps of the BVDR and CS methods for higher images, 20 and 16 iterations for lower ones.

We will use for initiating of the iterative optimization procedures the images given by convolution of the initial images with IPSF's. The relaxation parameter $\delta t$ of the schemas will be restricted with the help of (92). As it follows from the expression (77) the stable dynamic of the regularization parameter is conditioned by the same relation. The iterative processes can be bounded by the following conditions: the first is

$$\left\langle \left\| S^{(k+1)} - S^{(k)} \right\|^2 \right\rangle = \varepsilon^{(k)} \leq 10^{-8} \tag{94}$$

that means convergence limit, the second condition – $\varepsilon^{(k+1)} > \varepsilon^{(k)}$, that means passing of a local minimum, the third condition is the restriction of the iterations number, in this case 20 and 10 ones for the BVDR and CS schemas. The CS schema was used in relaxation free mode – $\delta t = 1$.

The results of image in figure 2 the second column of the third row optimization by the BVDR and CS schemas are shown in figure 5. There were used optimized IPSF's (44) and (64). This example shows that optimized by the procedure (54) IPSF (44) and IPSF (64) are equivalent in actuation on the result. The results of the optimization of the images in figures 3 and 4 are shown in figure 6. The results have shown that the schemas (78) and (85) give different results. Their convergence process is different too.

The schema (78) has some transition period at the first iteration steps. Initially $\lambda$ (77) grows up to the some maximum value and then monotonically reduces. The residual in (85) repeats $\lambda$ dynamic with one step lag. The transition period duration amounts up to 10 iteration steps depending on image structure and relaxation parameter. If image has many sharp contours then period is long, if image is mainly smooth than the period amounts of 2-4 iteration steps. The transition period amounts up to 20 steps and more when relaxation parameter $\delta t > 0.1$. So, we used $\delta t = 0.1$. The minimal relaxation parameter (92) $\delta t \sim 0.06$. The transition period did not exceed 5 iteration steps and convergence process was terminated by iteration steps number restriction. The limit residual was in vicinity (94). The convergence of the schema (78) is conditioned by $\lambda^{(k+1)} \leq \lambda^{(k)}$ ($k \gg 1$) in (77) since the $l_1$ norm of smoothed variation in numerator of expression (77) first fraction is less then the same norm of the excited variation of the function in the denominator. The reducing of the regularization parameter causes decrease of the residual in (94). The transition period presence is caused by initial difference between $X$ and $S^{(0)}$ and by dynamic nature of the regularization parameter (77).

The convergence of the schema (85) is monotonous but not unlimited primary. It reaches some lowest point of the residual in (94) and then the convergence condition (76) is violated. The point of convergence termination in accordance with (91) relates with image structure, filters order and relaxation parameter. The iteration steps number did not exceed 10 ones at $\delta t = 1.0$. But estimated image can be too mach sharpen. The images in figure 6 was given by no more 20 iterations at $\delta t = 0.1$, the minimal relaxation period (92) $\delta t_{min} \sim 0.07...0.08$. If $\delta t < \delta t_{min}$ then schema (85) is not convergent.

The presented results show that the optimization procedures sharpen and smooth image signal in relation with its structure. Images given by the CS schema are more smoothed in comparison with images given by BVDR schema at same relaxation parameter and iterations number. Small details in the figures show that the method CS is more precise in comparison with the BVDR method.

The parameter of convergence velocity $\theta$ did not exceed 1.0001 in (76) for both schemas (78) and (85).

The SAF (46) provides better convergence of iterative schema (78) in comparison with TV functional (67) and TV with $\beta$ - factor functional [15]. The schema (78) with mentioned regularization functionals is stable and convergent at $\delta t \ll 0.1$. Therefore, these functionals do not allow change primary image estimate by some iteration steps essentially.

*4.4. Deblurring of noisy images*

Iterative schemas [36] and (78), (85) are not appropriate for estimation of blurred and noised images because they sharpen all image surface curves and so reinforce noise influence, especially, when noise is impulsive. Noised and blurred image is presented in figure 6 high row at the left. Two PSF estimates by the methods [36] and (41) are similar and differ by phase of the maximums and fluctuations. They show that deblurring with the help of such PSF is impossible. The impulse noise correlation suppresses image correlation and therefore the PSF estimates correspond to noise feature. The CNS has been created by 59 eigenvector that does not correspond to image blur level. It is necessary to use a prior image denoised estimate $X_{pr}$ instead $X$ in [36] and (78), (85).

Denoised image estimate can be obtained by filtering with regularization in spectral domain [5, 12, 30, 41, 44] or by iterative schemas as special procedure [8, 15, 27, 28, 29, 39, 49] or as procedure joined with deconvolution [3, 4, 27, 50]. The filtration based approach introduces an additional smoothing or blur. The iterative schemas need a large number of iteration steps because usually use small value of relaxation parameter.

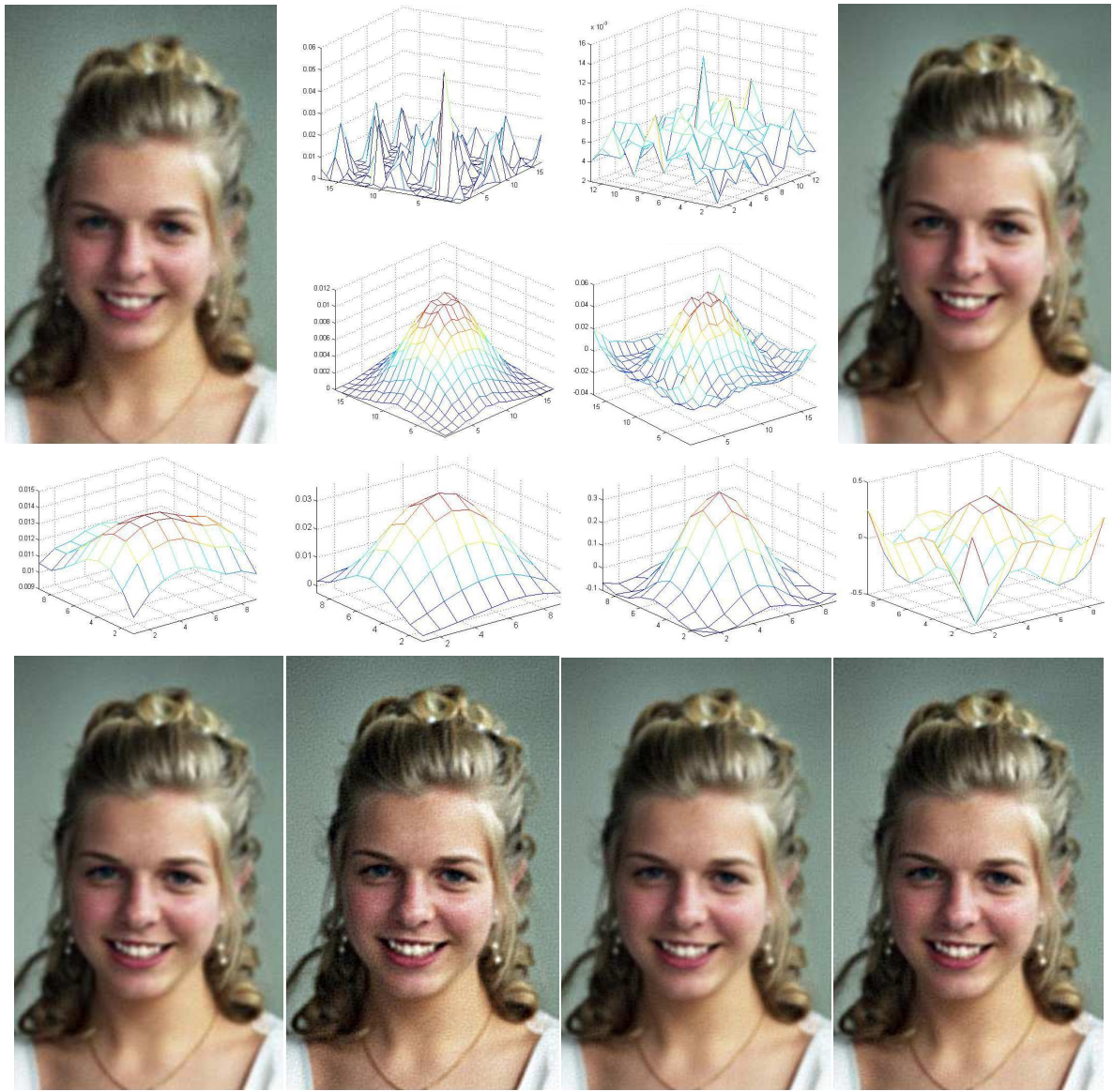

**Figure 7.** The noised and blurred image (AI=0.0206651), PSF estimates by the methods [36] and (41) (high sub-row), PSF and transient response of the denoising filter (sub-row below), denoised image at the right (AI=0.041148) are presented in the high row. The [36] PSF, the CNS method optimized by (50) PSF and IPSF's given by (54), (64) are shown in the second row. The image estimates by the method [36] (AI=0.033744), by the convolution with IPSF's (54) – AI=0.045611, and (64) – AI=0.041848, by 8 iterations of the CS method – AI=0.042248, are presented in the lower row.

The filter (56) with transient characteristic which is defined by equation (44) can be used as denoising filter for prior image $X_{pr}$ estimation because such filter does not introduce additional smoothing. If the filter order is high then it can eliminates noise influence by weighted accumulation of image matrix elements. The filter has to be corresponding to maximal blur level and therefore its CNS is given by single eigenvector of smallest eigenvalue.

The PSF defined by single eigenvector in (42) is presented in the lower sub-row of figure 7 first row. Its inverse form $G_{pr}$ is the transient characteristic of sought denoising filter. It was defined using equation (58), it is shown beside. The model parameters are the following: $P \times Q = 33 \times 33$, $L \times M = 17 \times 17$. The denoised image $X_{pr} = G_{pr} * X$ is shown in figure 7 high row at the right. It is denoised and partly deblurred. So, the given filter works as a linear bilateral filter.

The filtered image was used for PSF's estimation of the size $9 \times 9$ by the method [36] and by the CNS method with the model parameters: $P \times Q = 25 \times 25$, $L \times M = 9 \times 9$. In this case the CNS was created by 3 eigenvectors. These PSF's and corresponding to them IPSF's are presented in figure 7 second row. Both PSF estimates point on blur which has horizontal elongation.

The images which were obtained by convolution with IPSF's are presented in figure 7 low row. The result of the optimization by CS method 8 iteration steps is presented at the right. The relaxation parameter $\delta t = 0.1$, $\delta t_{min} \approx 0.05$. The result of the BVDR method is similar to it. The AI's analysis shows that estimated by CNS method blur characteristic is most exact. The optimization procedures smooth previous image estimates in accordance with regularization restriction as well as in the previous examples.

**Conclusion**

A singular matrix operator NS includes coupled left- and right-side vector spaces. Objects presented in the operator definition space can be projected into one of the NS. The projection operation splits the objects. The splitting relates with operator structure and the spaces size. This fact is the base of the presented approach to blind PSF estimation. Similar approaches were considered in [1, 33, 45]. Initial image polynomial model, which can be considering as 2D autoregression, was used for creation of the matrix operator which left side NS – conjugated NS, is the sought PSF space. The space dimension relates with blur level in wide diapason – from $\delta$-function in a trivial case to convex function with elongated or symmetrical shape which corresponds to blur feature. The PSF as a unique vector in CNS can be defined by his spectrum in the basis of CNS eigenvectors. The fact that the CNS is the PSF space is confirmed indirectly by the IPSF definition in the same space. The experiments have shown that this definition of the IPSF as a solution of ill-posed problem is most exact.

The CNS method is appropriate to PSF estimation when blur is characterizing by a smooth convex function. The spread function of image sharp displacement is not available to estimate.

Two methods of the IPSF estimation as transient characteristic of filter (2) are offered. The optimization of its shape allows compensate the PSF estimate inaccuracy and reconstruct blurred image by single convolution procedure with quality comparable with known iterative methods. The method in the CNS spectral space is more exact in comparison with space method when CNS is matched in vector space of the dimension $K \geq 3$.

If blur is homogeneous over all image surface then small fragment can be used for evaluation PSF and IPSF by the CNS method. So, the complexity of the method does not dependent from an image size.

Two approaches to image estimate optimization with additional restriction given by image minimal surface square as regularization term and as induced metric of the image space are presented. The CS schema (85) is most appropriate to optimization of primary estimate in real time implementation. The required iteration steps for reaching of the functional (83) local minimum does not exceed 20 ones at $\delta t = 0.1$. But only metric tensor determinant is applicable for image regularization in the case of the CS schema. At the same time, all known regularization functionals can be used in the BVDR schema (77)-(78) with the aim of restriction of image gradient or fluctuation, noise elimination etc.

The IPSF and image estimate optimization allows eliminate the PSF inaccuracy. Numerical experiments have shown that appropriate quality of blurred image reconstruction by optimization procedures can be reached with using approximate PSF that is given by single CNS eigenvector in (41). So, we have presented the solution of the deblurring problem in large when limit result does not strongly relate from results of separate phases.